\documentclass[10pt,twocolumn,letterpaper]{article}
\usepackage[pagenumbers]{cvpr}
\usepackage{microtype}
\usepackage{balance}
\usepackage{booktabs}
\usepackage{multirow}
\usepackage{pifont}
\usepackage{tikz}
\usetikzlibrary{positioning,arrows.meta,fit,calc}
\usepackage{listings}
\definecolor{cvprblue}{rgb}{0.21,0.49,0.74}
\usepackage[breaklinks,colorlinks,allcolors=cvprblue]{hyperref}

\newcommand{\method}{\textsc{ArtStage}\xspace}
\newcommand{\R}{\mathbb{R}}
\newcommand{\staging}{\mathbf{x}}
\newcommand{\labels}{\mathbf{z}}

\newcommand{\papertitle}{From Story to Staging: 3D Figures, Camera, and Light References \\ Learned from Film}
\title{\papertitle}
\author{Yunge Wen\\
{\normalsize Massachusetts Institute of Technology}\\
{\tt\small yungew@mit.edu}
\and
George Xi Wang\\
{\normalsize Stony Brook University}\\
{\tt\small george.x.wang@stonybrook.edu}
\and
Chengze Li\\
{\normalsize Saint Francis University}\\
{\tt\small czli@sfu.edu.hk}}
\date{}

\hypersetup{pdftitle={From Story to Staging: 3D Figures, Camera, and Light References Learned from Film},pdfauthor={Yunge Wen, George Xi Wang, Chengze Li}}
\begin{document}
\raggedbottom

\twocolumn[{%
\renewcommand\twocolumn[1][]{#1}%
\maketitle
\begin{center}
  \includegraphics[width=\textwidth]{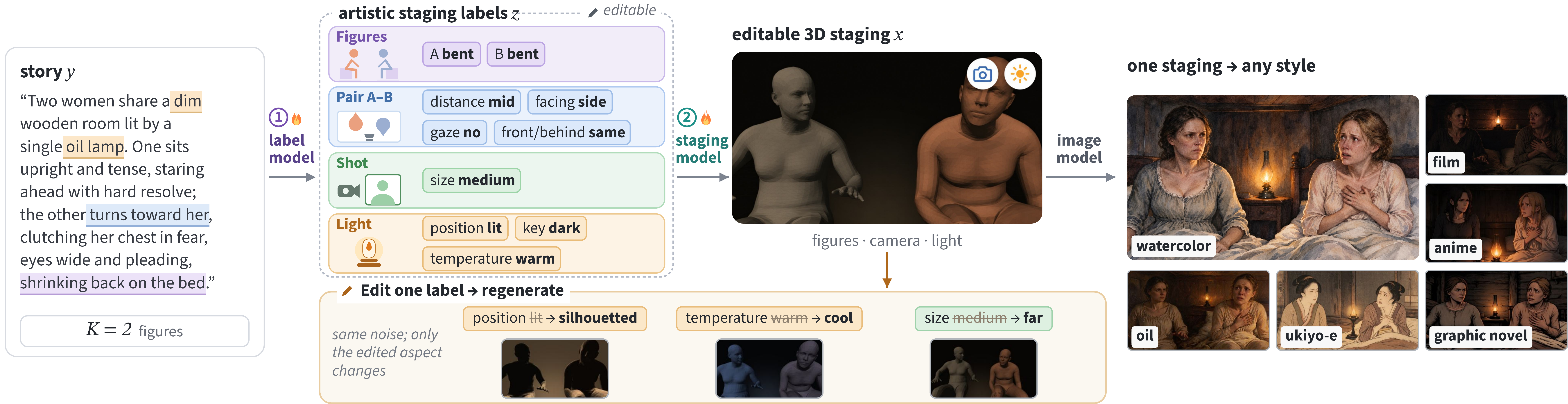}
  \captionof{figure}{\textbf{From story to staging.} From a short story and the number of figures $K$, model~\ding{172} predicts interpretable staging labels and model~\ding{173} turns them into posed 3D bodies, a camera, and lighting cues. The staging guides an image model in any style; any label can be edited and the staging regenerated from the same noise.}
  \label{fig:teaser}
\end{center}
}]

\begin{abstract}
Composing multi-person scenes requires translating narrative relationships, actions, and emotions into spatial arrangements, poses, camera viewpoints, and lighting. Existing generation methods either offer limited geometric control or require artists to specify layouts manually. We study \emph{text-to-staging}, the task of generating editable 3D compositions directly from short narrative briefs. We present \method, a two-stage framework that connects narrative understanding with geometric generation through interpretable staging labels. A text model predicts individual poses, pairwise relationships, camera framing, and lighting labels. A conditional flow-matching model then generates SMPL bodies, camera parameters, and lighting descriptors. We train both stages on 67,679 film stills with reconstructed stagings and narrative briefs. On held-out test stills, \method achieves 62.4\% label agreement, outperforming direct text-to-3D generation (57.8\%) and retrieval (50.4\%). Its outputs are preferred by a vision-language judge and nine expert illustrators. The labels also enable targeted edits while largely preserving the remaining composition.
\end{abstract}

% ===== sections/intro =====
% ===== Introduction =====
\section{Introduction}
\label{sec:intro}

% P1: What the artist needs.
Like a stage director, an illustrator must translate a narrative into a visual composition. Before drawing faces or costumes, the illustrator considers the relationships between characters, their interactions, and the emotions the scene should convey. These narrative elements inform concrete staging decisions: where each figure stands, how they are posed, whom they face or look at, their relative depth, the camera viewpoint, and the lighting. Together, these decisions define what we call a \emph{staging}. While experienced artists often make such choices intuitively, less experienced artists may find it difficult to translate a narrative into a coherent spatial arrangement.

% P2: why current tools fall short.
Existing tools make the process of translating a story into a concrete staging laborious. In 3D software, each figure's joints must be posed manually before the composition can be evaluated. Text-to-image models produce images quickly, but their output consists of fixed pixels, making it difficult to adjust figures and viewpoint independently. They also often fall back on a generic arrangement that ignores the relations in the story, as shown in our experiments (Sec.~\ref{sec:control-eval}).
Controllable generators accept additional conditions such as bounding boxes, keypoints, or 3D layouts~\cite{li2023gligen,zhang2023controlnet,wang2025cinemaster}, but users must create these conditions themselves, effectively performing the staging task in advance. Text-to-pose methods~\cite{delmas2022posescript,feng2024chatpose} generate a single body pose from a sentence describing that pose. More closely related methods jointly generate character and camera motion from text~\cite{cheng2026storytelling}, or editable object and camera trajectories~\cite{kizil2026lamp}. However, they focus on animation, with text specifying actions over time. To the best of our knowledge, no existing method takes a story describing a static scene and produces multiple posed figures and a viewpoint that express it.

% P3: the task.
In this work, we define this task as \emph{text-to-staging} and we propose \method to learn this task. Specifically, the input is the artist's story written down in a few sentences, which we call a \emph{narrative brief}. A brief says who the figures are to each other, what happens between them and what each one does, how the scene feels, and what light it is in (Fig.~\ref{fig:teaser}, leftmost). It comes with the number of figures $K\in\{1,\ldots,4\}$.
% OPEN (George): in Fig. 1 the story says "turns toward her" but the labels show "facing side, gaze no".
% narrative brief? or you just call it story?
% OPEN (George): Fig. 1 labels $y$ "story"; Fig. 2 and Sec. 4.1 call $y$ a "narrative brief". Use one label for $y$.
The output is a staging, which consists of $K$ posed SMPL bodies and one camera in a shared 3D frame, together with a coarse description of the light (Fig.~\ref{fig:teaser} middle). This description says whether the figures are lit or silhouetted against the background, how dark the picture is, and how warm the light is. The staging specifies no appearance; faces, costumes, and surroundings can subsequently be added by an artist or an image generator. Because a brief does not uniquely determine the arrangement, the same input should support multiple plausible stagings.

% P4: the key design choice: labels as an interpretable, editable intermediate.
% will you remove the branding later? Since you did not say anything about the branding at this moment. 
\method produces a staging in two steps, with a set of discrete \emph{staging labels} in between (Fig.~\ref{fig:teaser}, steps~\ding{172} and~\ding{173}). The labels describe a staging using established film and illustration concepts~\cite{mascelli1965fivecs,katz1991shot,block2008visual}, covering individual posture and isolation, pairwise spatial and interaction relationships, camera framing, and lighting attributes. Once a first model has predicted these labels from the brief, a second model generates the 3D staging from them, and running it again with different noise gives an alternative staging for the same labels. Because the labels read like ``she turns away from him'' or ``a closer shot'', an artist can change one and regenerate, which changes that relation in 3D while most of the rest of the staging stays in place (Sec.~\ref{sec:edit-eval}).

% P5: the data.
Training the two models requires stories that an expert has already turned into stagings, and film offers them at scale. In every shot, a director and a cinematographer have arranged the actors, the camera, and the light to tell one moment of a story, so learning from film shots means learning the expertise of directors. A film still, however, comes with neither its 3D staging nor its story in writing, and we recover both for 67{,}679 MovieNet stills~\cite{huang2020movienet} with one to four people (Sec.~\ref{sec:data}). We reconstruct the bodies and the camera with HMR2~\cite{goel2023hmr2}, measure the light on the frame, and derive the staging labels by fixed rules. A vision-language model that sees the frame then writes the briefs, which describe the relations, what each person does, the mood, and the light, but never the camera or where anyone is on screen, since these are what the model has to decide.

% P6: results and application.
As no earlier method produces a staging from a story, we compare \method with the same network conditioned directly on the brief, without labels, and with retrieval of the closest training shot. On held-out test stills, stagings generated through labels agree with the real shot on 62.4\% of the labels, against 57.8\% and 50.4\% for these two alternatives. Nine expert illustrators judging blind also prefer our stagings to both, and changing a single label produces the requested change in most cases while almost all other labels stay as they were (Sec.~\ref{sec:experiments}). Once the artist is satisfied with a staging, it can be opened in 3D software or passed to an image generator, which renders it in any style and reflects each label edit (Fig.~\ref{fig:teaser}, right and bottom). The artist thus directs the picture through decisions that are learned from directors' shots and stated in familiar terms, and the generated images follow the postures and the light of the staging more often than images made from the brief alone (Sec.~\ref{sec:control-eval}).
% OPEN (George): confirm the illustrator study (30 briefs, 540 judgments).
We summarize our contributions as follows.

\begin{itemize}
    \item We define text-to-staging: generating an editable 3D arrangement of posed figures and a camera, with a coarse description of the light, from a narrative brief and a figure count.
    \item We construct 135{,}344 brief--staging pairs from 67{,}679 film stills, with rule-based staging labels derived from reconstructed 3D geometry and lighting labels measured on the frame and validated against the briefs.
    \item We introduce \method, a two-stage model that predicts interpretable staging labels from text and generates 3D stagings conditioned on them, demonstrating that these labels improve agreement with the reference shot while enabling editing.
\end{itemize}

% ===== sections/relatedwork =====
% ===== Related Work =====
\section{Related Work}
\label{sec:related}

\paragraph{Language-to-layout planning.}
Several works turn language into a structured arrangement. ATISS and DiffuScene generate indoor object layouts with transformers and diffusion~\cite{paschalidou2021atiss,tang2024diffuscene}, and LayoutFlow applies flow matching to graphic layouts~\cite{guerreiro2024layoutflow}. LayoutGPT prompts a language model to write layouts directly~\cite{feng2023layoutgpt}, and LAMP has it write programs in a cinematography-inspired language that map deterministically to object and camera trajectories~\cite{kizil2026lamp}. These works arrange furniture, graphic elements, or moving objects rather than articulated bodies. Like LAMP, we plan through an interpretable intermediate, but our labels describe relations between figures, are learned from film rather than written by a language model, and condition a generative model instead of fixing the geometry; a language model predicting the same labels is a baseline (Sec.~\ref{sec:joint-eval}).

\paragraph{Human pose and camera.}
Language-guided human synthesis includes static pose prediction and temporal motion generation. PoseScript and ChatPose connect language to the 3D pose of a single body~\cite{delmas2022posescript,feng2024chatpose}, but neither places several people in relation to each other or to a viewer. Text-to-motion models such as MDM and T2M-GPT build on HumanML3D~\cite{guo2022humanml3d,tevet2023mdm,zhang2023t2mgpt}, InterGen and Shan et al.\ extend motion synthesis to interacting people~\cite{liang2024intergen,shan2024groupmotion}, and CCD and E.T.\ generate camera trajectories from text~\cite{jiang2024ccd,courant2024et}. Closest to us, Cheng et al.\ jointly synthesize two-person and camera motion from film~\cite{cheng2026storytelling}. These formulations prescribe actions or camera paths over time; \method generates a single moment of up to four people with its viewpoint and light from a description of their situation.

\paragraph{Controllable image generation.}
Image and video generators accept spatial conditions chosen by a user or another system: bounding boxes in GLIGEN, poses in ControlNet and HumanSD~\cite{li2023gligen,zhang2023controlnet,ju2023humansd}, loose depth and 3D boxes in LooseControl, and rendered SMPL maps in Champ~\cite{bhat2024loosecontrol,zhu2024champ}. CineMaster and CineScene control video with 3D boxes, camera trajectories, and scene features~\cite{wang2025cinemaster,huang2026cinescene}, and Frame In-N-Out extends motion control beyond the frame~\cite{wang2025frameinnout}. LightIt, DiLightNet, IC-Light, and VidCRAFT3 control lighting~\cite{kocsis2024lightit,zeng2024dilightnet,zhang2025iclight,zheng2026vidcraft3}. All of them assume that the poses, positions, viewpoint, and light have already been decided. \method makes these decisions from text, and its rendered staging can serve as the control for such generators.

\paragraph{Learning from film.}
Film and illustration manuals describe staging through shot size, blocking, and the visual structure of the frame~\cite{mascelli1965fivecs,katz1991shot,block2008visual}, and MovieNet annotates thousands of movies with shot- and scene-level information~\cite{huang2020movienet}. HMR2, which we use to recover the figures, reconstructs 3D people from single images such as film frames~\cite{goel2023hmr2}, and film footage has supervised joint human and camera motion~\cite{cheng2026storytelling}. For artworks, Human-Art and ArtEmis annotate people and emotional responses~\cite{ju2023humanart,achlioptas2021artemis,mohamed2022artemis2}, and Jen{\'\i}{\v{c}}ek and Chum retrieve paintings by pose~\cite{jenicek2019linking}. These works analyze or retrieve shots; we use single film shots, with their recovered 3D and measured light, as supervision for generating new stagings from text.

% ===== sections/data =====
% ===== Film-Derived Data =====

\begin{figure}[t]
    \centering
    \includegraphics[width=\linewidth]{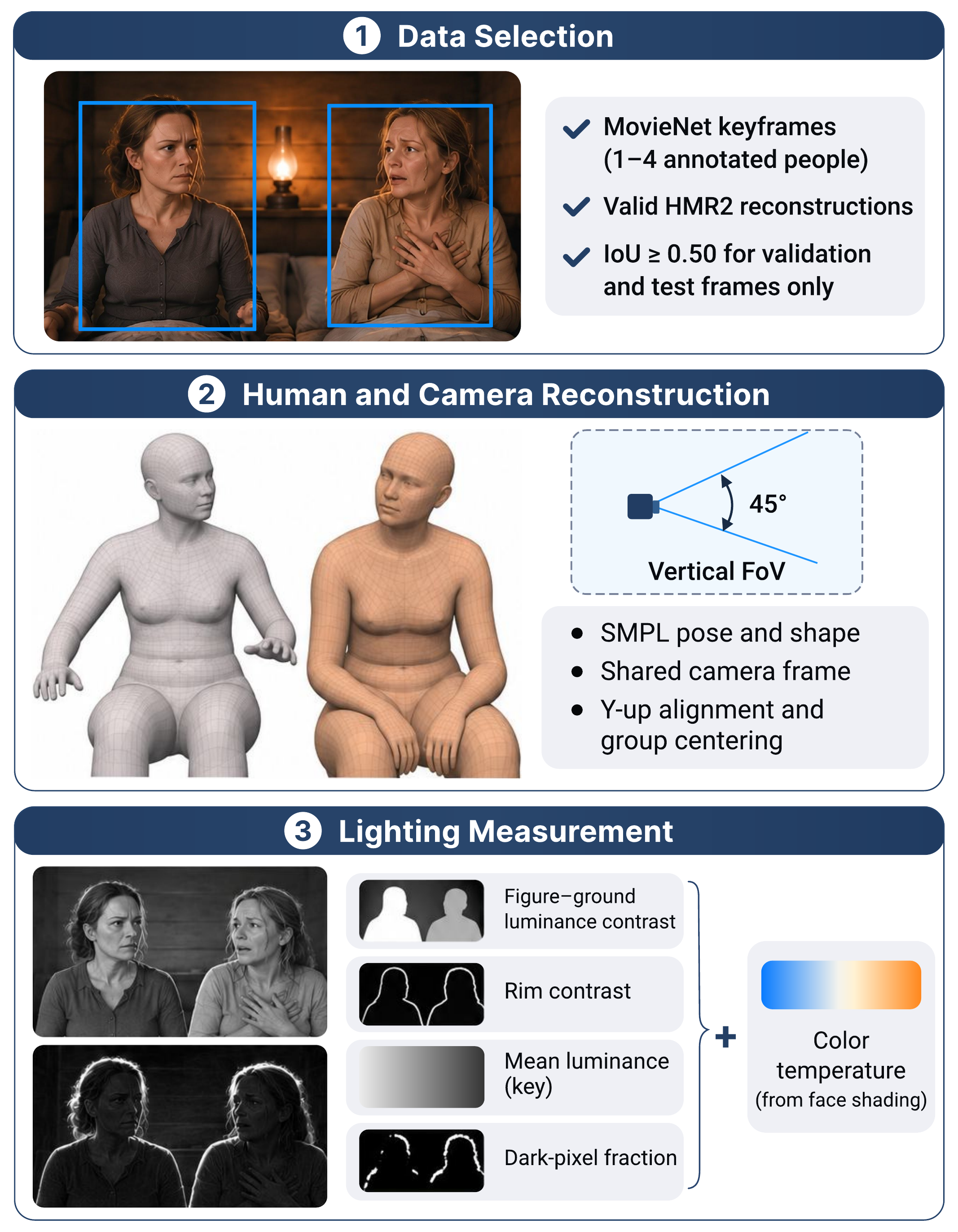}
    \caption{\textbf{Film-still staging recovery.}
    (1) We select MovieNet frames with one to four annotated people and valid HMR2 reconstructions.
    (2) We recover SMPL bodies and the camera in a shared coordinate system.
    (3) We measure the lighting and estimate its color temperature.}
    \label{fig:data_pipeline}
\end{figure}

\section{Dataset}
\label{sec:data}

We construct paired narrative briefs and 3D stagings from film stills. For each still, we recover the figures and camera, measure coarse lighting (Fig.~\ref{fig:data_pipeline}), derive semantic labels, and write narrative briefs. Figure~\ref{fig:qual} shows stills with their briefs and recovered stagings.

\subsection{Recovering Staging from Film Stills}
\label{sec:recovery}

\paragraph{Sources and selection.}
We use MovieNet keyframes~\cite{huang2020movienet} in which one to four people are annotated with bounding boxes, and the number of these people gives $K$. From each box, HMR2~\cite{goel2023hmr2} estimates the person's SMPL~\cite{loper2015smpl} pose, shape, and camera translation. We discard stills with missing or invalid reconstructions, and for validation and test stills we additionally require an IoU of at least $0.50$ between each projected body and its annotated box.

\paragraph{Human and camera reconstruction.}
We convert crop-level translations into a shared camera coordinate system, assuming a $45^\circ$ vertical field of view, then transform the recovered bodies into a $Y$-up world, align the ground plane, and center the group. The camera is level, retains the original shot direction, and is parameterized relative to the mean body root (Sec.~\ref{sec:representation}). Rendering the stored bodies from this camera reproduces the composition of the still, as the columns \emph{Film still} and \emph{Real shot} of Fig.~\ref{fig:qual} show.

\label{sec:light}
\paragraph{Lighting measurement.}
A directional light fitted to the shading of the faces does not agree with the light described in the briefs. We therefore describe the light by four quantities measured directly on the pixels, namely the figure--ground luminance contrast, the rim contrast, the mean luminance (key), and the dark-pixel fraction. The fifth quantity, the color temperature, is taken from the face-shading fit, whose light color does agree with warm and cold descriptions. 

\subsection{Semantic Staging Labels}
\label{sec:labels}

Following film and illustration terminology~\cite{mascelli1965fivecs,katz1991shot,block2008visual}, we derive discrete labels from the recovered staging, covering individual posture and isolation, pairwise distance, facing, gaze, and front/behind order, camera shot size, and lighting attributes. These labels are semantic in that each one names what a story decides, such as who looks at whom or who stands apart, and not the joint angles that realize it. People are indexed left to right on screen. Labels are computed by deterministic rules using body joints, camera depth, and lighting measurements, with categories defined by geometric thresholds or training-set tertiles. Three professional artists reviewed the vocabulary and definitions. A vision-language check showed weaker agreement for facing and gaze than for other labels.

\subsection{Narrative Briefs}
\label{sec:briefs}

We give the vision-language model DeepSeek-V4.1-Flash each still together with its MovieNet captions and ask it to write two briefs of at most 50 words, describing relationships, actions, attention, emotion, and lighting without explicit camera, framing, left/right, or depth terms. Because the model sees each still, the briefs may contain visually inferred relationships, and we assess how informative they are about the labels in Sec.~\ref{sec:label-eval}. The resulting 135{,}344 briefs (up to two per still) cover 67{,}679 stills. We use 66{,}522 / 576 / 581 stills for training, validation, and testing, excluding validation and test shots from training.

% ===== sections/method =====
% ===== Method =====
\begin{figure*}[t]
  \centering
  \includegraphics[width=\textwidth]{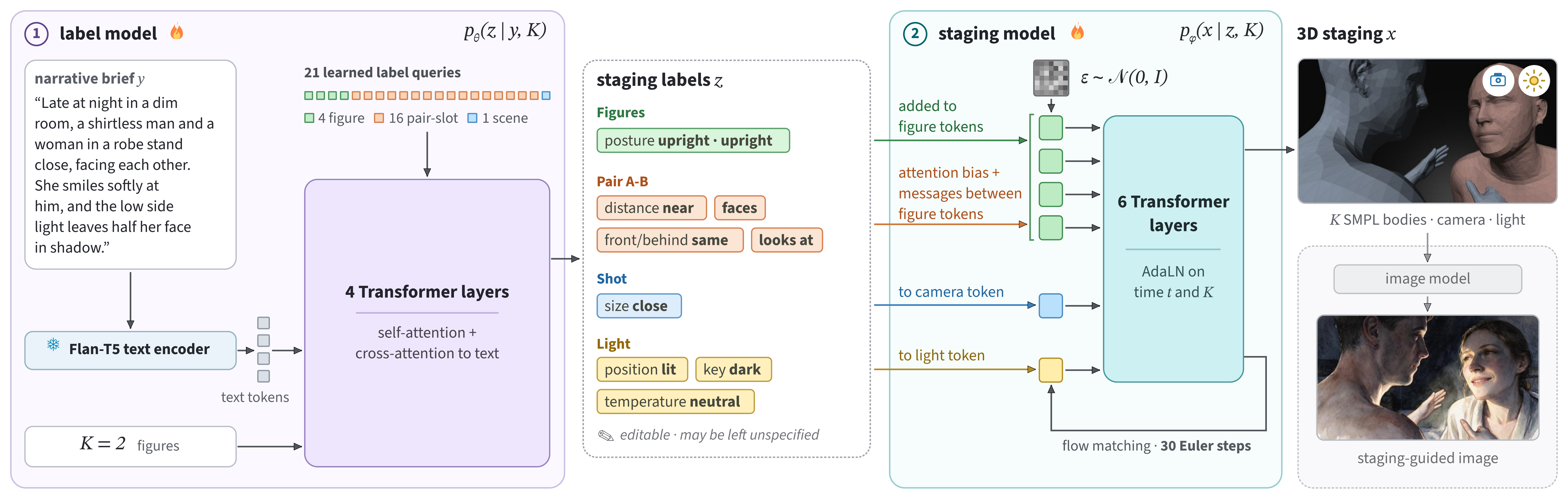}
  \caption{\textbf{\method.} \ding{172} A label model reads the brief with a frozen text encoder and predicts every staging label with one query per figure, ordered pair, and scene. \ding{173} A flow-matching Transformer turns Gaussian noise into figure, camera, and light parameters, conditioned on the labels, which enter the tokens and attention as shown. Labels can be edited or left unspecified before \ding{173} is run.}
  \label{fig:arch}
\end{figure*}

\section{Method}
\label{sec:method}

\method follows the two-stage factorization of Eq.~\eqref{eq:factor} (Fig.~\ref{fig:arch}). A label model predicts the staging labels from the brief (Sec.~\ref{sec:t2l}); a flow-matching model generates the 3D staging and the light from the labels (Sec.~\ref{sec:l2s}). The two models are trained separately.

% ===== sections/problem =====
\subsection{Task Formulation and Staging Representation}
\label{sec:task}

The input is a narrative brief $y$, a short text about who the figures are to each other, what is happening between them, and how it feels, together with the number of figures $K$. The output is a staging
\begin{equation}
    \staging = \bigl(\{p_i\}_{i=1}^{K},\,c,\,\ell\bigr),
    \qquad 1 \leq K \leq 4,
    \label{eq:staging}
\end{equation}
where $p_i$ specifies the pose, body shape, and position of figure $i$, $c$ the camera, all in one 3D frame, and $\ell$ a coarse description of the light seen from that camera.

\label{sec:representation}
Each figure is represented as
\begin{equation}
    p_i=[\rho_i,\beta_i,q_i]\in\R^{157},
    \qquad \rho_i\in\R^{24\times6},\ \beta_i\in\R^{10},\ q_i\in\R^3,
\end{equation}
with 6D joint rotations~\cite{zhou2019rotation} (including the global orientation), SMPL shape, and root translation in the $Y$-up world of Sec.~\ref{sec:recovery}. The camera retains the original shot direction and is represented relative to the mean body root as
\begin{equation}
    c=(\phi,\sin\psi,\cos\psi,\log r)\in\R^4,
    \label{eq:camera}
\end{equation}
where $r$, $\phi$, and $\psi$ denote camera distance, elevation, and azimuth. At decoding it is placed at $a+r(\cos\phi\sin\psi,\sin\phi,\cos\phi\cos\psi)$ and looks along $-z$ with a $45^\circ$ vertical field of view. Figure and camera parameters are standardized with training-split statistics. The light $\ell\in\R^5$ is the frame descriptor of Sec.~\ref{sec:light} (figure--ground contrast, rim, key, share of dark pixels, color temperature). Figures occupy four slots ordered from left to right on screen; unused slots are masked.

We use the staging labels $\labels=g(\staging)$ defined in Sec.~\ref{sec:labels}; any real or generated staging can thus be described and checked using the same deterministic labels. We model the conditional distribution of stagings through the labels,
\begin{equation}
    p(\staging\mid y,K)=\sum_{\labels} p_\theta(\labels\mid y,K)\;p_\phi(\staging\mid \labels,K),
    \label{eq:factor}
\end{equation}
and, at inference, take the most likely labels under $p_\theta$ and draw several stagings from $p_\phi$. An artist can also bypass $p_\theta$ and set, change, or leave unspecified any label.

\subsection{Narrative-to-Label Prediction}
\label{sec:t2l}

The brief is encoded once by a frozen Flan-T5-base encoder~\cite{chung2024flan} (at most 48 tokens). The label model holds one learned query per label slot: four figure queries, one query per slot of a $4\times4$ pair grid (the 12 off-diagonal slots are the ordered pairs; the diagonal is masked), and one scene query that predicts the shot size and the three light labels, each with learned slot embeddings and an embedding of $K$. Four Transformer layers (width 256, four heads) alternate self-attention among the queries with cross-attention to the text tokens, and a linear head per label type outputs class probabilities. We train with cross-entropy over the slots that exist for the given $K$. Gaze and set-apart are rare (most figures look at nobody and are not isolated), so these two label types are weighted by inverse class frequency. We train three seeds with early stopping on validation macro recall and keep the best. At inference we take the most likely class of each label.

\subsection{Relation-Conditioned Staging Generation}
\label{sec:l2s}

The staging model is a flow-matching Transformer~\cite{lipman2023flow,liu2022rectified}. Let $\bar\staging_1$ be a standardized training staging, $\epsilon$ Gaussian noise, and $t\sim\mathcal U(0,1)$. With $\bar\staging_t=(1-t)\epsilon+t\bar\staging_1$, the network $v_\phi(\bar\staging_t,t,\labels,K)$ regresses the velocity $u=\bar\staging_1-\epsilon$; the loss is a weighted mean-squared error over global orientation, body pose, shape, and position of the existing figures, plus the camera and the light.

\paragraph{Tokens.} Four figure tokens, one camera token, and one light token are linear projections of their parameters to width 384, plus type embeddings and, for figures, slot embeddings. Six Transformer blocks with adaptive LayerNorm conditioned on flow time and $K$ update the tokens.

\paragraph{Label conditioning.} Each label type has an embedding table with an extra entry for \emph{unspecified}. Posture and set-apart embeddings are added to the figure tokens, the shot-size embedding to the camera token, and the light-label embeddings to the light token. Pair labels condition the interaction between two figures: the summed embeddings of all pair labels of $(i,j)$ are projected to a per-head bias on the attention logit from $i$ to $j$ and to a message that $j$ sends to $i$. This allows pairwise labels to condition interactions between the corresponding figure tokens. A learned bias per token-type pair covers attention between figures, camera, and light.

\paragraph{Robustness to imperfect labels.} At test time the labels come from the label model and can be wrong, or an artist may specify only some of them. During training we therefore replace 10\% of the label entries by \emph{unspecified}, corrupt another 10\% to a random class, and drop all labels of 10\% of the examples. The last case also trains the unconditional model used for classifier-free guidance~\cite{ho2022cfg}.

\paragraph{Training and sampling.}
Both models are trained with AdamW; the staging model for 60 epochs with an exponential moving average (details in the appendix). We sample with 30 Euler steps and guidance weight $w=2$, $\tilde v=v_\varnothing+w(v_{\labels}-v_\varnothing)$; different noise gives alternative stagings for the same labels.

\subsection{Editable Staging Generation}
\label{sec:editing}

Any label can be set, changed, or left unspecified, and the staging can be regenerated from the same noise (Sec.~\ref{sec:edit-eval}). The generated staging is exported as SMPL meshes and camera parameters, while the image-based lighting descriptors are translated into a renderable light configuration. Specifically, a physical key light is placed on the camera side or behind the figures, with its intensity, exposure, and color adjusted according to the predicted lighting attributes. This controls the visual appearance of illumination without reconstructing the original scene's physical lighting. The rendered staging can then be used as a control for image generation (Sec.~\ref{sec:control-eval}).

% ===== sections/evaluation =====
% ===== Experiments =====
\section{Experiments}
\label{sec:experiments}

\begin{table*}[t]
  \centering
  \setlength{\tabcolsep}{3.5pt}
  \resizebox{\textwidth}{!}{%
  \begin{tabular}{@{}lc ccccccc ccc c cc c@{}}
    \toprule
    & \multicolumn{11}{c}{Label agreement} & & \multicolumn{2}{c}{Balanced} & \\
    \cmidrule(lr){2-12}\cmidrule(lr){14-15}
    Method & Mean [95\% CI] & post & apart & dist & face & look & depth & cam & pos & key & temp & look$^+$ & strict & matched [95\% CI] & contact \\
    \midrule
    Random (same $K$) & .461 [.454, .469] & .49 & .87 & .39 & .39 & .66 & .37 & .42 & .34 & .34 & .34 & .25 & .387 & .437 [.428, .447] & .27 \\
    Retrieval & .504 [.496, .515] & .54 & .87 & .47 & .41 & .66 & .43 & .53 & .39 & .40 & .36 & .23 & .433 & .487 [.479, .496] & .27 \\
    $K$ only $\to$ 3D & .464 [.456, .472] & .49 & .88 & .41 & .37 & .67 & .37 & .43 & .35 & .34 & .33 & .20 & .389 & .442 [.434, .452] & .25 \\
    Text $\to$ 3D & .578 [.564, .592] & .60 & .88 & .57 & .44 & .69 & .44 & .65 & .46 & .62 & .43 & .28 & .506 & .563 [.551, .575] & .23 \\
    $K$-prior labels $\to$ 3D & .511 [.499, .525] & .49 & \textbf{.91} & .44 & .42 & \textbf{.75} & .46 & .51 & .39 & .34 & .38 & .06 & .405 & .426 [.416, .437] & .33 \\
    LLM labels $\to$ 3D & .510 [.498, .521] & .59 & .87 & .42 & .43 & .63 & .46 & .38 & .34 & .55 & .43 & \textbf{.45} & .477 & .520 [.506, .534] & .55 \\
    \textbf{Text $\to$ labels $\to$ 3D (ours)} & \textbf{.624 [.611, .634]} & \textbf{.65} & .89 & \textbf{.63} & \textbf{.50} & .69 & \textbf{.51} & \textbf{.72} & \textbf{.53} & \textbf{.63} & \textbf{.49} & .38 & \textbf{.534} & \textbf{.570 [.559, .581]} & .23 \\
    \midrule
    True labels $\to$ 3D (upper bound) & .947 [.942, .951] & .97 & .97 & .91 & .85 & .84 & .94 & .97 & 1.00 & 1.00 & 1.00 & .74 & .924 & .930 [.922, .940] & .22 \\
    \bottomrule
  \end{tabular}}
  \caption{\textbf{Generated stagings compared with the real shot} (1{,}162 test briefs). We report label agreement per type and its mean, where pos, key, and temp are the light labels and look$^+$ is the recall of real gaze relations. Balanced agreement is given in screen order (strict) and after matching generated to real figures (matched). Contact is the share of multi-figure stagings with two figures' joints closer than 8\,cm (real shots: .19). Given the true labels, the staging model satisfies 94.7\% of them.}
  \label{tab:main}
\end{table*}

\begin{figure*}[t]
  \centering
  \includegraphics[width=\textwidth]{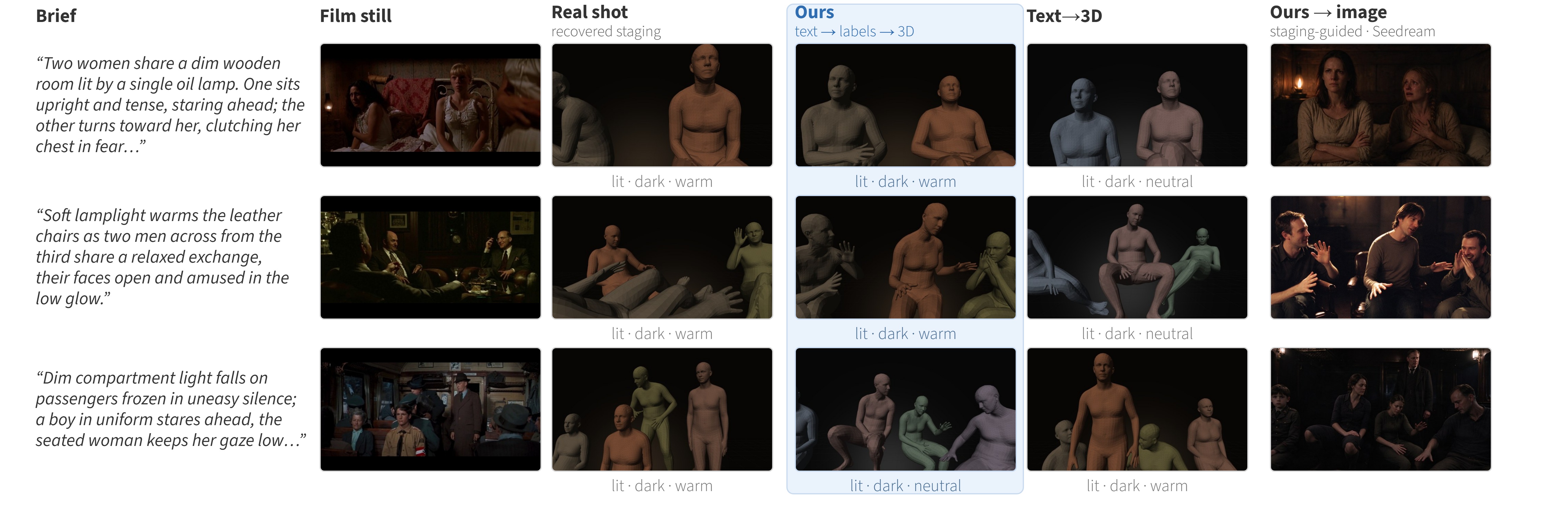}
  \caption{\textbf{Qualitative comparison on test briefs.}
We compare the reference film still, recovered staging (real shot), our method, and direct Text-to-3D generation (both best of four). Our method better preserves character poses, spatial relationships, and lighting. The rightmost column shows images generated from our stagings.}
  \label{fig:qual}
\end{figure*}

\subsection{Setup}
\label{sec:exp-setup}

\paragraph{Test data.} All results use the 1{,}162 briefs of the 581 test stills; neither these stills nor any other frame of their shots is used in training. Each method receives a brief and $K$ and returns one staging.

\paragraph{Metric.} We compare a generated staging with the real shot, and for the light the labels are computed from the generated light descriptor with the same cut points. \emph{Label agreement} is the fraction of agreeing label entries, computed per label type and averaged over the ten types. Because several label types are dominated by one class, we also report \emph{balanced} agreement (per-class recall averaged within and over types). We give it in screen order (\emph{strict}) and after matching generated to real figures with the best of the $K!$ assignments (\emph{matched}), since a brief rarely says who stands on the left. Confidence intervals are 95\% percentile intervals from 300 bootstrap resamples of test movies. We also report the share of multi-figure stagings in which joints of two different figures come within 8\,cm (\emph{contact}), which is $0.19$ for the real shots.

\paragraph{Baselines.} \emph{Random} draws a training staging with the same $K$. \emph{Retrieval} returns the staging of the training brief closest in pooled text-encoder features. \emph{$K$ only} is our staging model with all labels unspecified. \emph{Text$\to$3D} is the same staging network, including the light token, conditioned on the brief through cross-attention to the text tokens instead of on labels and trained identically, so it tests whether going through the labels helps. \emph{$K$-prior labels} feeds the most common class of every label slot given $K$ to our staging model. \emph{LLM labels} asks a large language model (DeepSeek-V4.1-Flash) to predict all labels from the brief, with two training examples in the prompt, and feeds them to our staging model. It returned usable labels for all 1{,}162 briefs. \emph{True labels} feeds the labels of the real shot and is an upper bound for the staging model.

\subsection{Do briefs determine the labels?}
\label{sec:label-eval}

Before any 3D staging is generated, the label model agrees with the real shot more often than the $K$-prior and the language model. It recovers 43\% of the real gaze relations and 95\% of the figures set apart. The briefs also carry lighting information: the label model predicts the key correctly in 63\% of cases and whether the figures are lit or silhouetted in 53\% (chance: 33\%).

\begin{figure*}[t]
  \centering
  \includegraphics[width=\textwidth]{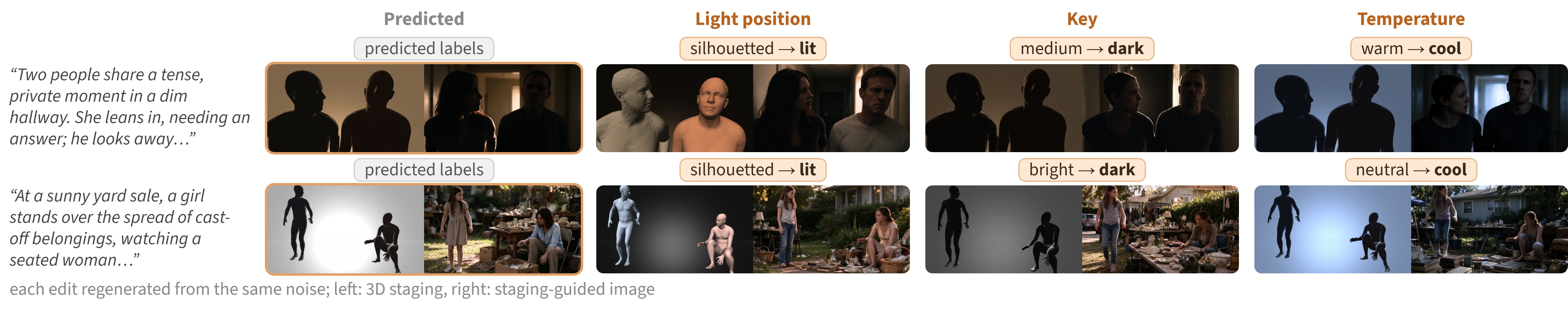}
  \caption{\textbf{Editing the light.} From the labels predicted for a brief, one light label is changed at a time and the staging is regenerated from the same noise (each pair: 3D staging, then the image generated with it). Poses, positions, and camera are kept; only the light changes.}
  \label{fig:edits}
\end{figure*}

\subsection{Generating the staging}
\label{sec:joint-eval}

Table~\ref{tab:main} gives the main comparison. Generating through labels agrees best with the real shot, with $.624$ against $.578$ for the same network conditioned directly on text and non-overlapping intervals. Retrieval reaches $.504$ and labels chosen by a language model reach $.510$, or $.565$ [.554, .575] with Claude Opus 5.5. Over the seven figure and camera labels alone the gap is larger ($.656$ against $.610$). The gain is spread over posture, distance, facing, front/behind, and shot size, and the labels also lead on all three light labels. Under balanced agreement the order is the same in screen order (strict, $.534$ vs.\ $.506$), and after matching figures our method remains first ($.570$ vs.\ $.563$). Figure~\ref{fig:qual} shows examples, with more in the appendix.

$K$-prior and language-model labels are often mutually inconsistent and yield many interpenetrating figures (contact $.33$ and $.55$), whereas our single samples come close to the real shots ($.23$ vs.\ $.19$). With the true labels, the staging model reaches $.947$ agreement, so the remaining gap lies in predicting labels from text, since briefs leave distance, facing, and front/behind largely open.

\paragraph{Judged against the story.} Label agreement compares with one real shot in our own label space. As a check that depends on neither the real shot nor our labels, a vision-language model (GPT-6.1 Terra) sees a brief and the two renders side by side (ours and Text$\to$3D, both best of four, sides balanced) and picks the staging that better shows the relations and mood. On 299 test briefs it prefers ours in 58.2\% of cases [52.5, 63.7] (95\% CI over movies), with no position bias (left chosen in 48.8\%). Nine expert illustrators, judging blind, preferred ours over Text$\to$3D and retrieval in 71.1\% [58.9, 81.5] and 72.6\% [59.6, 84.1] of 540 judgments ($p\le0.002$).

\paragraph{Robustness.} With the test briefs rewritten from the frames alone by a different model (Claude Opus 5.5, no captions), ours still leads ($.620$ [.608, .629] vs.\ $.581$ [.568, .592] for Text$\to$3D). On unseen InterHuman~\cite{liang2024intergen} mocap (1,176 test sequences, human-written descriptions), ours cuts the body-facing and head-direction errors of Text$\to$3D from $49.0^\circ$/$46.6^\circ$ to $43.2^\circ$/$39.6^\circ$ ($p<0.001$).

\paragraph{Choosing among samples.} Since every staging can be checked against the predicted labels, we draw four stagings and keep the one that satisfies most predicted labels without interpenetration. This lowers contact from $.23$ to $.06$ and raises label following from $.92$ to $.94$ at unchanged agreement ($.626$ [.613, .637]; matched $.571$).

\paragraph{Measures independent of the labels.} We also compare the stagings with the real shots directly in 3D: pose Fr\'echet distance, Wasserstein distance of eight layout quantities, interpenetration, floating, and figures outside the frame. At guidance weight 1, our best-of-four stagings have the poses closest to the real shots among the generative methods (Fr\'echet distance $.03$ vs.\ $.04$ for Text$\to$3D), floating and off-screen rates at the level of the real shots, and keep their lead in label agreement ($.621$ vs.\ $.551$). Lowering the weight from 2 to 1 brings the layout closer to the real shots (Wasserstein distance $.33\to.15$) at almost unchanged agreement ($.626\to.621$). The front/behind comparison keeps its order for thresholds from 0.2 to 0.5\,m.

\subsection{Editing through labels}
\label{sec:edit-eval}
\begin{table}[t]
  \centering
  \small
  \setlength{\tabcolsep}{4pt}
  \resizebox{\linewidth}{!}{%
  \begin{tabular}{@{}lrcccc@{}}
    \toprule
    Edited label & $n$ & Success & Before & Kept & Kept (resample) \\
    \midrule
    Posture & 153 & .58 & .01 & .99 & .96 \\
    Distance & 83 & .70 & .04 & .95 & .91 \\
    Facing & 102 & .51 & .08 & .95 & .91 \\
    Gaze & 100 & .53 & .15 & .98 & .95 \\
    Front/behind & 84 & .71 & .04 & .92 & .92 \\
    Shot size & 156 & .76 & .01 & .97 & .95 \\
    Light: position & 150 & .99 & .00 & .99 & .93 \\
    Light: key & 163 & 1.00 & .00 & .99 & .95 \\
    Light: temperature & 171 & .99 & .00 & .99 & .94 \\
    \midrule
    Figure and camera labels & 678 & .63 & .05 & .96 & .94 \\
    Light labels & 484 & .99 & .00 & .99 & .94 \\
    \bottomrule
  \end{tabular}}
  \caption{\textbf{Single-label edits.} \emph{Success} means that the edited staging has the requested value, and \emph{Before} that the unedited staging already had it. \emph{Kept} is the share of the other labels unchanged by the edit, and \emph{Kept (resample)} is the same share between two samples of the unedited labels.}
  \label{tab:edit}
\end{table}

For every test brief we take the labels predicted by the label model, change exactly one label to a different class (a figure's posture, a pair's distance, facing, gaze, or front/behind, the shot size, or one of the light labels), and regenerate from the same noise. For symmetric relations we change both directions. An edit succeeds if the regenerated staging has the requested label. We also measure how many of the other labels stay unchanged, and compare with how many stay unchanged between two samples of the unedited labels with different noise.

Table~\ref{tab:edit} reports figure and camera edits and light edits separately. Edits of the figures and the camera produce the requested change in 63\% of cases, against 5\% in which the unedited staging already had the requested value, while facing and gaze, which require a coordinated change of many joints, succeed about half the time. Light edits succeed in 99\% of cases, which is expected rather than evidence of physical lighting, because the light labels are tertiles of the generated light descriptor and the light token is conditioned on them directly. The edits are also local, in that 96--99\% of the other labels are unchanged, more than the 94\% that survive a change of noise, and a light edit moves the joints by 13\,cm on average against 68\,cm for a new sample.
Figure~\ref{fig:edits} shows light edits, each of which changes the light as requested while the figures and camera stay in place.

\subsection{Staging as a control for image generation}
\label{sec:control-eval}

We give an image generator (Seedream 5.0 Flash~\cite{seedream2025}) the brief with no control, 2D boxes, 2D skeletons, or our rendered staging. Without the staging it falls back on a generic symmetric arrangement; with it, gesture, depth order, and framing carry over (examples in the appendix). A vision-language model reading the images finds that those made from our staging follow its posture (69\% vs.\ 49\%), figure--ground light (69\% vs.\ 47\%), and light color (72\% vs.\ 57\%) more often than those made from the brief alone (all measures in the appendix).

% ===== sections/discussion =====

% ===== Discussion =====
\section{Discussion}
\label{sec:discussion}

\paragraph{Where the remaining gap lies.}
Given the true labels, the staging model satisfies 94.7\% of them; given labels predicted from the brief, the staging agrees with the real shot on 62.4\%. The bottleneck is therefore reading the story, not building the 3D. Much of this gap is not an error in the usual sense: a brief rarely fixes how far apart two people stand or who is closer to the viewer, and the director's shot is one of several stagings that would serve the same story. Agreement with a single shot is thus a conservative measure, which is why we also ask a judge and illustrators whether a staging fits the brief. Because the two stages are trained separately, a better label predictor, or an artist who fills in the labels the story leaves open, improves the result without retraining the staging model.

\paragraph{Labels as an interface.}
The labels separate what the story decides (who looks at whom, who stands apart, how close the shot is) from what the data decides (the joint angles that realize it). This is what makes edits local: one changed label leaves more of the staging intact than a new sample does. The interface is only as good as its weakest labels. Some relational labels are less reliable and harder to edit, particularly when changes require coordinated adjustments across multiple joints. Light is the opposite case: it follows its labels almost perfectly because it is a description of the frame rather than a light in the scene.

\paragraph{Limitations.}
Stagings are recovered from single frames with a fixed $45^\circ$ field of view and an untilted camera, and cover at most four figures. The guidance weight $w=2$ was not tuned for editing. The light is a coarse frame descriptor, not a physical light: it does not say from which side the light comes, and a bright background can pass for back light. The labels are coarse rules; gaze and facing, estimated from single-frame body and head orientation, are not confirmed by an independent reading of the stills. Edits that need coordinated changes of many joints succeed about half the time. The briefs come from one vision-language model that has seen the frame.

\paragraph{Data and broader impact.}
The training frames are copyrighted film stills; the model learns from geometry, light measures, and text derived from them and outputs only bodies, a camera, and a light descriptor, not frames. Its stagings inherit the conventions of the films in MovieNet, for instance who tends to be staged as dominant, isolated, or in shadow, as well as the biases of HMR2 and of the model that wrote the briefs.

\paragraph{Future work.}
Natural extensions are a physical light with a direction, scenes with props and more than four people, a field of view and camera tilt learned from data, and stagings that unfold over several shots or in time, where the same labels could serve as keyframes for animation and video generators.

% ===== Conclusion =====

\section{Conclusion}
\label{sec:conclusion}

We introduced text-to-staging and \method, a framework that generates editable 3D arrangements of human figures, camera, and lighting from narrative briefs. By using interpretable staging labels to connect narrative intent with geometry, our approach improves agreement with reference stagings over direct text-to-3D generation. Expert evaluations further support its ability to convey the intended narrative. The labels also enable targeted edits while largely preserving the remaining composition. Our results demonstrate the potential of semantic staging representations for controllable visual composition and artistic creation.

\clearpage

\balance
\section*{LLM Usage Statement}
Large language models were used in three roles in this work. (1) As part of the method and data: DeepSeek-V4.1-Flash, a vision-language model, wrote the narrative briefs from the film stills and their captions; it also served as the language-model baseline that predicts staging labels and as the automatic reader of the generated images. GPT-6.1 Terra, a vision-language model from a different family, served as the automatic judge of stagings. Claude Opus 5.5, a model from a third family, served as a second language-model baseline and rewrote the test briefs from the frames alone, without captions, as a robustness check. All prompts are given in the appendix. Seedream 5.0 Flash, an image generation model, produced the images guided by our stagings. (2) As a coding assistant, an LLM helped write and debug data-processing, training, and evaluation scripts; all code was run, checked, and is the responsibility of the authors. (3) As a writing assistant, an LLM helped draft and edit the text of this paper; the authors reviewed all content and take full responsibility for it, including the reported numbers, which come from our experiments.

{\small
\bibliographystyle{ieeenat_fullname}
\bibliography{main}
}

\clearpage
\nobalance
% Appendix included by main.tex.
\appendix
\counterwithin{figure}{section}
\counterwithin{table}{section}
\counterwithin{equation}{section}
\twocolumn[{\centering\Large\bfseries Appendix\par\vspace{0.6em}}]
\lstset{
  basicstyle=\ttfamily\scriptsize,
  breaklines=true,
  breakatwhitespace=false,
  columns=fullflexible,
  keepspaces=true,
  linewidth=\linewidth,
  frame=single,
  framesep=2pt,
  xleftmargin=4pt,
  xrightmargin=4pt,
  aboveskip=4pt,
  belowskip=6pt,
  rulecolor=\color{black!30}
}
\noindent This appendix provides additional details: data recovery and light measurement (Appendix~\ref{sec:s-data}), the prompts (Appendix~\ref{sec:s-prompts}), model and training details (Appendix~\ref{sec:s-model}), the rendering of the light (Appendix~\ref{sec:s-render}), additional results (Appendix~\ref{sec:s-results}), and the interactive editor (Appendix~\ref{sec:editor}).

%======================================================================
\section{Data}
\label{sec:s-data}

\begin{figure}[t]
  \centering
  \includegraphics[width=\linewidth]{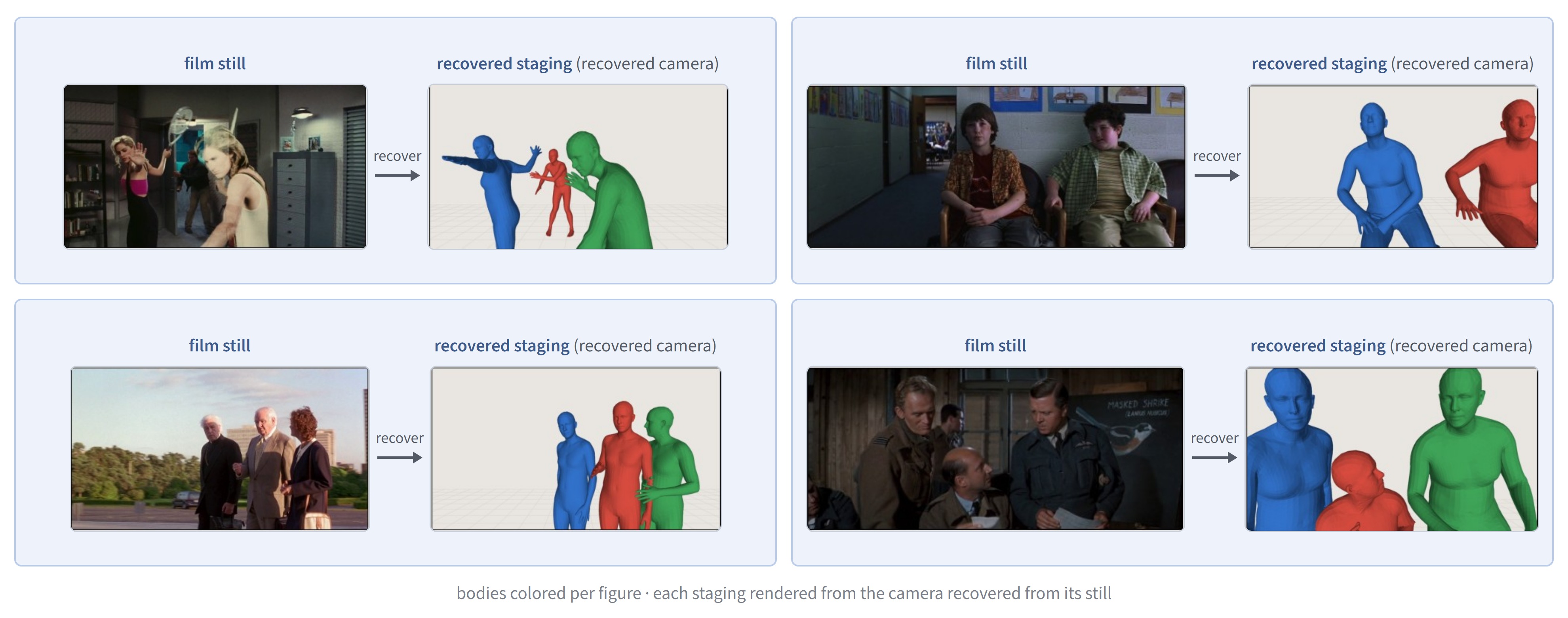}
  \caption{\textbf{Recovered stagings.} Left: film stills. Right: the staging recovered from each still, rendered from the recovered camera.}
  \label{fig:s-recon}
\end{figure}

\paragraph{Recovering the staging.}
For each annotated person box in a MovieNet keyframe, HMR2 predicts SMPL pose and shape and a weak-perspective camera for the crop. Crop translations are converted to a common full-image camera with a focal length corresponding to a $45^\circ$ vertical field of view. A frame is kept if the number of reconstructed people equals the number of annotated boxes; for the validation and test stills, every body projected into the frame must also overlap its box with IoU $\geq 0.50$. Points are moved into a $Y$-up world by the rotation $\mathrm{diag}(1,-1,-1)$; the same rotation must be applied to each body's global orientation, which HMR2 predicts in the camera frame. HMR2 places the SMPL mesh with its rest-pose pelvis $J_0(\beta)$ at the predicted translation, so the mesh is offset by $J_0(\beta)$ (rotated into the world) from the stored root. With both details handled, a stored staging rendered from its camera reproduces the reconstruction (Fig.~\ref{fig:s-recon}).

\begin{figure*}[t]
  \centering
  \includegraphics[width=\textwidth]{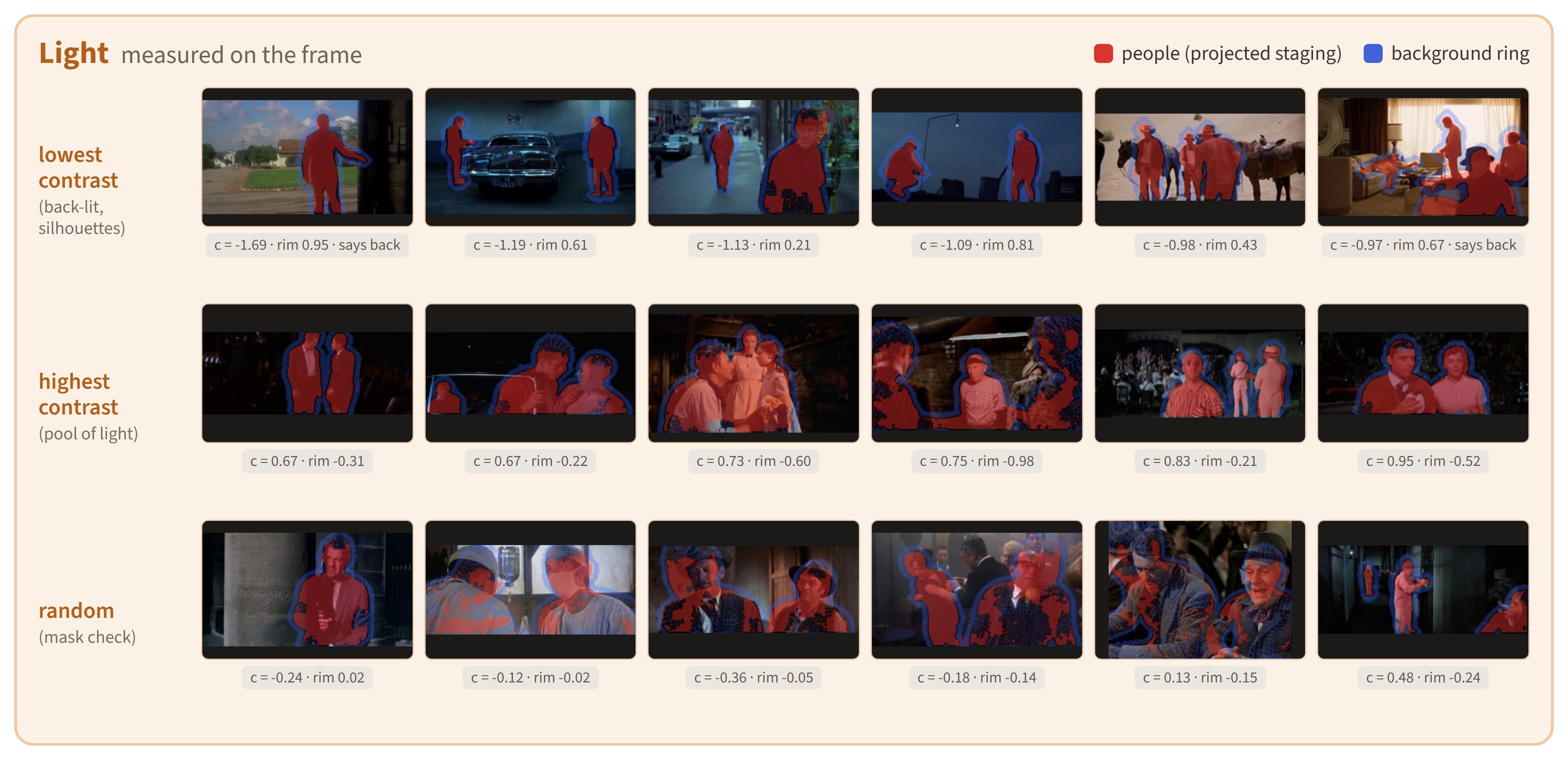}
  \caption{\textbf{Light measured on the frame.} Red: people projected from the recovered staging; blue: the background ring. Top: lowest figure--ground contrast (silhouettes, back light; the desert scene shows that a bright ground also counts). Middle: highest contrast (people in a pool of light). Bottom: random stills, showing that the projected bodies cover the people. $c$: contrast, rim: rim measure.}
  \label{fig:s-light}
\end{figure*}

\paragraph{Light measured on the frame.}
We project all SMPL vertices of the recovered bodies into the frame (resized to a width of 320 pixels), mark the hit pixels, close the mask with two iterations of binary closing, and fill holes. On the luminance $Y=0.2126R+0.7152G+0.0722B$ we compute, with $\epsilon=0.02$,
\begin{itemize}
  \item figure--ground contrast $\log\frac{\bar Y_{\text{people}}+\epsilon}{\bar Y_{\text{ring}}+\epsilon}$, where the ring is the mask dilated by 10 pixels minus the mask dilated by 2 pixels;
  \item rim $\log\frac{\bar Y_{\text{outline}}+\epsilon}{\bar Y_{\text{core}}+\epsilon}$, with the outline the mask minus its one-pixel erosion and the core its three-pixel erosion;
  \item key, the mean luminance of the frame, and the share of pixels with $Y<0.1$.
\end{itemize}
Stills whose mask covers less than 1\% or more than 80\% of the frame get no contrast or rim (0.6\% of the training stills lack a measurement and take the training median). The color temperature is taken from the face-shading fit of the extraction: per color channel, the coefficient of the first-order spherical-harmonics term along the fitted light direction gives a light color, and $t=0.5+0.7\,(b-r)/(b+r)$, clipped to $[0,1]$, places neutral light near $0.5$, warm light below, and cold light above. Figure~\ref{fig:s-light} shows the masks and the stills with the most extreme contrast. Light labels are tertiles of the contrast (silhouetted / even / lit), key (dark / medium / bright), and temperature (warm / neutral / cool) on the training split.
The tertile cut points on the training split are $-0.168$ / $0.094$ for the contrast, $0.145$ / $0.239$ for the key, and $0.196$ / $0.443$ for the color temperature.

\paragraph{Staging label vocabulary.}
Seven geometric label types are computed from SMPL forward kinematics and camera parameters; the three lighting label types are defined by the measurements above. Invalid figure slots and self-pairs are masked. Figures are ordered left to right on screen. The geometric rules are:
\begin{itemize}
  \item \textbf{Posture:} lying if the angle between the pelvis-to-neck vector and the world up direction exceeds $60^\circ$; otherwise bent if the mean of the left and right hip--knee--ankle angles is below $130^\circ$; upright otherwise.
  \item \textbf{Set apart:} for $K\geq3$, a figure is set apart if its nearest-neighbor horizontal pelvis distance is maximal among the figures and exceeds 1.5 times the median nearest-neighbor distance. For $K<3$, no figure is marked set apart.
  \item \textbf{Distance:} horizontal pelvis-to-pelvis distance, discretized into near, mid, and far using training-set tertiles over unordered valid pairs, with cut points at $0.6000$ and $1.0490$\,m.
  \item \textbf{Facing:} angle between the horizontal projection of the first figure's root forward axis and the horizontal direction to the second pelvis: toward below $60^\circ$, away above $120^\circ$, side otherwise.
  \item \textbf{Gaze:} yes if the angle between the first figure's head forward axis and the 3D direction from its head to the second head is below $35^\circ$; no otherwise. This measures head orientation, not eye gaze.
  \item \textbf{Front/behind:} signed camera-depth difference $d_i-d_j$: in front below $-0.3$ m, behind above $0.3$ m, same otherwise.
  \item \textbf{Shot size:} close, medium, and far are the training-set tertiles of the log camera radius $\log r$, with cut points at $0.0024$ and $0.2894$.
\end{itemize}
Distance is symmetric; facing, gaze, and depth are ordered-pair quantities. The remaining three labels are the lighting categories defined above. Three professional artists reviewed the staging label vocabulary and definitions, as reported in the main paper.

\paragraph{Dataset splits.}
The dataset contains 67,679 stills and 135,344 briefs (up to two per still). The training, validation, and test splits contain 66,522, 576, and 581 stills, respectively. Validation and test shots are excluded from training. The test split contains 1,162 briefs.

\paragraph{Validation against the briefs.}
We count a brief as describing a kind of light with the following regular expressions:
\par
\begingroup
\ifdefined\nolinenumbers\nolinenumbers\fi
\begin{lstlisting}
back:   silhouett|backlit|back-lit|against the (bright|glare|window|light|door|doorway|sun|sky|glow)|light (from )?behind
front:  (face|faces) (lit|illuminated|bathed|catch)|light (falls|falling|spills) (on|across|over) (his|her|their)
dark:   dim|dimly|dark|darkness|night|gloom|gloomy|shadowy|candlelit|candlelight|moonlit|murky|low light
bright: bright|brightly|sunlit|sunny|daylight|noon|glaring|sun-drenched|well-lit|brilliant
\end{lstlisting}
\par
\endgroup
 Among stills whose briefs describe back light, 61\% fall in the lowest contrast tertile; among those describing lit faces, 43\% fall in the highest (33\% by chance). The face fit separates these groups barely better than chance (AUC .54, .53, .47; Table~\ref{tab:light}); its color temperature agrees with warm and cold descriptions (51\% and 48\% in the matching tertile).

\begin{table}[t]
  \centering
  \small
  \setlength{\tabcolsep}{3pt}
  \resizebox{\linewidth}{!}{%
  \begin{tabular}{@{}llcc@{}}
    \toprule
    Briefs say & $n$ & Frame measure & Face fit \\
    \midrule
    silhouetted, back-lit & 226 & \textbf{.68} (contrast) & .54 \\
    faces lit & 296 & \textbf{.59} (contrast) & .53 \\
    dim, night vs.\ bright, sunlit & 2{,}731 / 1{,}854 & \textbf{.92} (key) & .47 \\
    \bottomrule
  \end{tabular}}
  \caption{\textbf{Light measures against the briefs.} AUC for separating the stills whose briefs describe a kind of light from the others (0.5: no relation). Frame measures: figure--ground contrast, mean luminance. Face fit: direction (front to back) and strength of the light fitted to the shading of the faces.}
  \label{tab:light}
\end{table}

%======================================================================
\section{Prompts}
\label{sec:s-prompts}

\par\medskip\noindent\textbf{Narrative briefs.}\par\nobreak
DeepSeek-V4.1-Flash receives the frame, the MovieNet captions, the number of people, and two worked examples, with the system prompt:
\par
\begingroup
\ifdefined\nolinenumbers\nolinenumbers\fi
\begin{lstlisting}
Rewrite visual descriptions of a film still (and the frame, if given) as the brief a director would write before staging it: who the people are to each other, what happens between them, the emotional tone.
- Say what each person does with body and attention: who watches whom, who turns away or avoids someone's eyes, who sits, kneels or lies down, who is left out or stands apart, who is close or touching.
- Say what light the scene is in, as part of the story: its source and quality and how it falls on the people (candlelight, harsh noon sun, cold office light, dusk through a window, faces half in shadow, silhouetted against a bright doorway). Read it from the frame.
- Generic people (a man, two sisters), no names.
- No camera, shot size, framing, angle, left/right, foreground/background.
- Keep the given number of people; do not contradict the description or the frame.
- Each brief at most 50 words, English.
Return JSON only: {"briefs": [...]}
\end{lstlisting}
\par
\endgroup

\par\medskip\noindent\textbf{LLM labels (baseline).}\par\nobreak
The language-model baseline receives two training briefs with their true labels as examples and the system prompt below. The prompt uses front/side/back for the figure--ground lighting categories lit/even/silhouetted, respectively; these are appearance categories rather than physical light directions.
\par
\begingroup
\ifdefined\nolinenumbers\nolinenumbers\fi
\begin{lstlisting}
Plan the staging of a film still from a brief. K people, numbered 0..K-1 left to right on screen. Return JSON only:
{"post": [per person: "upright"|"bent"|"lying"],   // bent = sitting, kneeling, crouching
 "apart": [per person: true if set apart from the others],
 "cam": "close"|"medium"|"far",
 "ldir": "front"|"side"|"back",   // front = people lit brighter than their surroundings; back = people darker than what is behind them (silhouette, back light)
 "lint": "dark"|"medium"|"bright",  // overall brightness of the picture (dark = low key, night)
 "ltemp": "warm"|"neutral"|"cool",
 "pairs": [every ordered pair i!=j: {"i", "j",
   "dist": "near"|"mid"|"far",
   "face": "toward"|"side"|"away",   // i's body toward j
   "look": true|false,               // i looks at j
   "depth": "in_front"|"same"|"behind"}]}   // i relative to j, seen from the camera
\end{lstlisting}
\par
\endgroup

\par\medskip\noindent\textbf{Staging preference judge.}\par\nobreak

\par
\begingroup
\ifdefined\nolinenumbers\nolinenumbers\fi
\begin{lstlisting}
You judge 3D stagings for a picture. A staging shows grey mannequin figures, a camera view and a light; it has no faces, clothes or setting on purpose. The image shows two stagings side by side, A (left) and B (right). Choose the staging that better shows the story: who the people are to each other, what is happening between them, and the mood (poses, positions, whom each faces or looks at, distances, framing, light). Ignore rendering quality. Answer JSON only: {"choice": "A" or "B", "reason": "<12 words"}
\end{lstlisting}
\par
\endgroup

\par\medskip\noindent\textbf{Generated image reading.}\par\nobreak

\par
\begingroup
\ifdefined\nolinenumbers\nolinenumbers\fi
\begin{lstlisting}
You read a picture of a scene with people. Number the people 0, 1, ... from left to right in the picture. Answer JSON only: {"n_people": int, "posture": [per person: "standing"|"sitting"|"lying"], "looks_at": [[i, j], ...] for every person i who looks at person j, "faces": [[i, j], ...] for every person i whose body is turned toward person j, "brightness": "dark"|"medium"|"bright", "figure_light": "lit" (people brighter than their surroundings) | "even" | "silhouette" (people darker than what is behind them), "warmth": "warm"|"neutral"|"cool"}
\end{lstlisting}
\par
\endgroup

\par\medskip\noindent\textbf{Film-still label validation.}\par\nobreak

\par
\begingroup
\ifdefined\nolinenumbers\nolinenumbers\fi
\begin{lstlisting}
You read a film still with people. Number the people 0, 1, ... from left to right in the picture. Answer JSON only: {"n_people": int, "posture": [per person: "standing"|"sitting"|"kneeling"|"crouching"|"lying"], "looks_at": [[i, j], ...] every person i who looks at person j, "faces": [[i, j], ...] every person i whose body is turned toward person j, "in_front": [[i, j], ...] every person i clearly closer to the camera than person j, "shot": "close"|"medium"|"far" (close: head and shoulders; far: full bodies small in frame), "brightness": "dark"|"medium"|"bright", "figure_light": "lit" (people brighter than their surroundings) | "even" | "silhouette" (people darker than what is behind them), "warmth": "warm"|"neutral"|"cool"}
\end{lstlisting}
\par
\endgroup

\par\medskip\noindent\textbf{Image generation.}\par\nobreak
The following templates are instantiated with the brief, style, figure count, and (for our staging) lighting words. The example below uses two figures; the code uses singular wording for one figure.
\par\medskip\noindent\textbf{Brief only.}\par\nobreak

\par
\begingroup
\ifdefined\nolinenumbers\nolinenumbers\fi
\begin{lstlisting}
[BRIEF] Paint this scene as [STYLE]. Show exactly [K] people. No text.
\end{lstlisting}
\par
\endgroup

\par\medskip\noindent\textbf{2D boxes.}\par\nobreak

\par
\begingroup
\ifdefined\nolinenumbers\nolinenumbers\fi
\begin{lstlisting}
[BRIEF] Paint this scene as [STYLE]. The reference image is only a layout guide with one coloured box per person: show exactly [K] people, each person filling one box at the same position and size, with the same camera framing. Do not draw the boxes. No text.
\end{lstlisting}
\par
\endgroup

\par\medskip\noindent\textbf{2D skeletons.}\par\nobreak

\par
\begingroup
\ifdefined\nolinenumbers\nolinenumbers\fi
\begin{lstlisting}
[BRIEF] Paint this scene as [STYLE]. The reference image is an OpenPose skeleton per person: show exactly [K] people with these exact body poses, positions and facing directions, and the same camera framing. Do not draw the skeletons. No text.
\end{lstlisting}
\par
\endgroup

\par\medskip\noindent\textbf{Rendered staging.}\par\nobreak

\par
\begingroup
\ifdefined\nolinenumbers\nolinenumbers\fi
\begin{lstlisting}
[BRIEF] Paint this scene as [STYLE]. The reference image is a staging guide: keep exactly [K] people with the same body poses, positions, distances, facing directions and gaze, the same camera viewpoint and framing, and the same lighting ([LIGHT WORDS]). Replace the grey mannequins with real characters, costumes and a setting that fit the story. No text, no grid lines.
\end{lstlisting}
\par
\endgroup

%======================================================================
\section{Model and training details}
\label{sec:s-model}

Table~\ref{tab:s-hparams} lists the hyperparameters of both models. The label model reads the frozen Flan-T5-base token features through four Transformer decoder layers whose 21 queries (4 figures, a $4\times4$ grid of pair slots whose 12 off-diagonal entries are the ordered pairs, 1 scene) cross-attend to the text; one linear head per label type produces the class logits, and pair outputs for distance are symmetrized. The staging model is a 6-layer Transformer over 4 figure tokens, 1 camera token, and 1 light token with adaptive LayerNorm on the flow time and $K$; pair labels enter every layer as per-head attention biases and as messages between the two figures.

\begin{table*}[t]
  \centering
  \small
  \setlength{\tabcolsep}{4pt}
  \begin{tabular}{@{}lll@{}}
    \toprule
    & Label model & Staging model \\
    \midrule
    Text encoder & Flan-T5-base, frozen, 48 tokens & -- (Text$\to$3D: same) \\
    Width / layers / heads & 256 / 4 / 4 & 384 / 6 / 6 \\
    Tokens & 4 figure, $4{\times}4$ pair, 1 scene queries & 4 figure, 1 camera, 1 light \\
    Dropout & 0.1 & 0.1 \\
    Optimizer & AdamW, lr $3{\times}10^{-4}$, wd 0.05 & AdamW, lr $2{\times}10^{-4}$, wd 0.01, $\beta=(0.9,0.95)$ \\
    Batch / epochs & 256 / $\leq$20, patience 4 & 256 / 60, cosine schedule \\
    Other & 3 seeds, select on val.\ macro recall & EMA 0.999, grad.\ clip 1.0 \\
    Loss & CE, inverse-frequency weights (gaze, set apart) & flow matching; weights 1/1/0.5/1 (orient./pose/shape/pos.) \\
    Label dropout & -- & 10\% unspecified, 10\% corrupted, 10\% all dropped \\
    Sampling & argmax & 30 Euler steps, guidance $w=2$ \\
    \bottomrule
  \end{tabular}
  \caption{\textbf{Hyperparameters.} Text$\to$3D uses the staging model with cross-attention to the text tokens, the pooled text feature added to the conditioning, and 10\% text dropout for guidance. The checkpoint with the lowest validation loss (evaluated every 5 epochs) is used.}
  \label{tab:s-hparams}
\end{table*}

The person loss is the sum of the four weighted blockwise mean squared velocity errors, averaged over valid figures. The camera and light losses are mean squared velocity errors over their respective components, each added with weight 1. Thus each block is normalized by its dimensionality rather than weighting all 157 person coordinates equally.

%======================================================================
\section{Rendering the light}
\label{sec:s-render}

The light descriptor is not a physical light rig; for figures we render it by convention, from its three labels. \emph{Position}: \emph{lit} places a key light on the camera side, above and to the side, over a dark surround with a soft pool of light around the group; \emph{even} uses soft light from above and the front over a mid-grey surround; \emph{silhouetted} places the key light behind the group, in front of a bright glow, with a rim term on the outlines. \emph{Key} scales the exposure (0.5, 0.9, 1.2), and \emph{temperature} sets the light color (warm, neutral, cool). The experimental figures shown in this paper use grey Lambertian meshes, lightly tinted per figure. The updated editor uses a common neutral, near-white base color for all figures and the same lighting convention. Its background is a separate image-space gradient centered on the projected group anchor, rather than a lit physical dome. The live viewport and exported camera image share the renderer's per-face colors, background profile, exposure, and light color; the rasterization implementations differ. Light direction is a visualization convention, not an additional predicted physical parameter.

%======================================================================
\section{Additional results}
\label{sec:s-results}

\paragraph{Labels predicted from briefs.}
Table~\ref{tab:labels} compares the label sources of Sec.~\ref{sec:label-eval} of the main paper before any 3D is generated.

\begin{table}[h]
  \centering
  \setlength{\tabcolsep}{2.2pt}
  \resizebox{\linewidth}{!}{%
  \begin{tabular}{@{}lcccccccccccc@{}}
    \toprule
    & & \multicolumn{7}{c}{Figures and camera} & \multicolumn{3}{c}{Light} & \\
    \cmidrule(lr){3-9}\cmidrule(lr){10-12}
    Labels from & Mean & post & apart & dist & face & look & depth & cam & pos & key & temp & look$^+$/apart$^+$ \\
    \midrule
    $K$-prior & .515 & .49 & \textbf{.92} & .44 & .43 & \textbf{.78} & .46 & .52 & .39 & .34 & .38 & .00 / .00 \\
    LLM (2-shot) & .489 & .59 & .75 & .41 & .41 & .59 & .46 & .36 & .34 & .55 & .43 & \textbf{.52} / .44 \\
    Label model (ours) & \textbf{.613} & \textbf{.65} & .73 & \textbf{.64} & \textbf{.51} & .71 & \textbf{.51} & \textbf{.73} & \textbf{.53} & \textbf{.63} & \textbf{.49} & .43 / \textbf{.95} \\
    \bottomrule
  \end{tabular}}
  \caption{\textbf{Labels predicted from briefs}, compared with the labels of the real shot (test split, before generating 3D). Columns are label agreement per type; light: figure--ground position (lit / even / silhouetted), key, and color temperature. look$^+$ and apart$^+$ are the recall of the rare classes ``looks at'' and ``set apart''. The class weighting of the label model trades agreement on these two types for recall of the rare class.}
  \label{tab:labels}
\end{table}

\paragraph{Light from the face fit vs.\ from the frame.}
Table~\ref{tab:s-light} compares the same pipeline trained with light labels from the face-shading fit (front/side/back of the fitted direction, fitted strength, temperature) and from the frame measures. With the face fit, position and strength agreement remain close to chance even when predicted by the language model; with the frame measures, both position and key become predictable from the briefs.

\begin{table}[h]
  \centering
  \small
  \setlength{\tabcolsep}{4pt}
  \begin{tabular}{@{}llccc@{}}
    \toprule
    Light labels from & Method & position & key / strength & temp. \\
    \midrule
    \multirow{3}{*}{face fit} & LLM labels & .33 & .33 & .45 \\
     & Text$\to$3D & .40 & .35 & .44 \\
     & Ours & .34 & .36 & .45 \\
    \midrule
    \multirow{3}{*}{frame} & LLM labels & .36 & .57 & .44 \\
     & Text$\to$3D & .46 & .59 & .43 \\
     & Ours & \textbf{.48} & \textbf{.60} & \textbf{.45} \\
    \bottomrule
  \end{tabular}
  \caption{\textbf{Light label agreement} of generated stagings with the real shot (three classes, chance .33). Both blocks come from a separate pair of training runs that differ only in the source of the light labels, so the frame block differs slightly from Table~\ref{tab:main} of the main paper.}
  \label{tab:s-light}
\end{table}

\paragraph{Choosing among samples.}
Table~\ref{tab:s-bo4} gives the full numbers for best-of-four sampling.

\begin{table}[h]
  \centering
  \small
  \setlength{\tabcolsep}{3pt}
  \resizebox{\linewidth}{!}{%
  \begin{tabular}{@{}lccccc@{}}
    \toprule
    Method & $w$ & Agreement [95\% CI] & Balanced matched & Follow & Contact \\
    \midrule
    Ours, one sample & 2 & .624 [.611, .634] & .570 & .923 & .234 \\
    Ours, best of 4 & 2 & \textbf{.626} [.613, .637] & \textbf{.571} & \textbf{.938} & .063 \\
    Ours, one sample & 1 & .610 [.598, .621] & -- & .861 & .217 \\
    Ours, best of 4 & 1 & .621 [.608, .632] & .567 & .907 & .025 \\
    Text$\to$3D, one sample & 2 & .578 [.564, .592] & .563 & -- & .226 \\
    Text$\to$3D, best of 4 & 2 & .579 [.566, .591] & .563 & -- & .033 \\
    Text$\to$3D, one sample & 1 & .554 [.541, .565] & -- & -- & .203 \\
    Text$\to$3D, best of 4 & 1 & .551 [.540, .561] & .530 & -- & .011 \\
    \bottomrule
  \end{tabular}}
  \caption{\textbf{Best of four samples.} Ours maximizes predicted-label agreement minus a unit penalty for contact; Text$\to$3D selects the first contact-free sample, or the first sample if none is contact-free. Contact is the joint-distance proxy defined in the main paper, not a direct mesh-interpenetration test. Real shots: contact .194.}
  \label{tab:s-bo4}
\end{table}

\paragraph{Diversity and locality.}
For each test brief we draw four samples and compare all pairs (Table~\ref{tab:s-div}). With the labels fixed, our samples keep the relations (6.2\% of labels differ) while the figure roots still move by 47\,cm and the camera varies; Text$\to$3D changes 34.0\% of the labels between samples, indicating greater variation in the measured staging attributes. With the same noise, editing one light label moves the joints by 13\,cm and barely changes the camera; editing the shot size changes the camera as intended.

\begin{table}[h]
  \centering
  \small
  \setlength{\tabcolsep}{3pt}
  \resizebox{\linewidth}{!}{%
  \begin{tabular}{@{}lccc@{}}
    \toprule
    & Labels differ & Root shift (m) & $|\Delta\log r|$ \\
    \midrule
    Ours, 4 samples & .062 & .472 & .187 \\
    Text$\to$3D, 4 samples & .340 & .444 & .178 \\
    \midrule
    & \multicolumn{2}{c}{Joint displacement (cm)} & $|\Delta\log r|$ \\
    \midrule
    Edit one light label & \multicolumn{2}{c}{12.6} & .045 \\
    Edit shot size & \multicolumn{2}{c}{24.1} & .419 \\
    New sample (re-sample) & \multicolumn{2}{c}{68.2} & .188 \\
    \bottomrule
  \end{tabular}}
  \caption{\textbf{Diversity and locality.} Top: mean over pairs of four samples per test brief. Bottom: change from our sample when one label is edited with the same noise, vs.\ drawing a new sample.}
  \label{tab:s-div}
\end{table}

\paragraph{Measures independent of the labels.}
Table~\ref{tab:s-3d} compares the generated stagings of the 1{,}162 test briefs with the real shots the briefs were written from. Pose FD is the Fr\'echet distance between the 23 local joint rotations (6D) of generated and real figures. Layout W1 is the 1-D Wasserstein distance to the real shots, in units of the real standard deviation, averaged over pair distance, front/behind offset, facing angle, gaze angle, knee angle, torso tilt, camera distance, and camera elevation. Contact as in the main paper; feet: for stagings with two or more standing figures, the spread of the lowest foot joint over the standing figures (median, and share above 25\,cm); off-screen: share of figures whose pelvis projects outside a 16:9 frame. Depth: front/behind agreement with the real shot when the threshold of the front/behind label is 0.2, 0.3 (used throughout), or 0.5\,m.

\begin{table*}[t]
  \centering
  \small
  \setlength{\tabcolsep}{4pt}
  \resizebox{\linewidth}{!}{%
  \begin{tabular}{@{}llccccccccc@{}}
    \toprule
    Method & $w$ & Pose FD & Layout W1 & Contact & Feet (m) & Feet $>$25\,cm & Off-screen pelvis & Depth@0.2 & Depth@0.3 & Depth@0.5 \\
    \midrule
    Real shots & -- & 0 & 0 & .194 & .131 & .21 & .375 & -- & -- & -- \\
    Retrieval & -- & .05 & .168 & .265 & .114 & .16 & .503 & .389 & .430 & .518 \\
    $K$-prior labels $\to$ 3D & 2 & 1.49 & .576 & .327 & -- & -- & .364 & .381 & .456 & .573 \\
    LLM labels $\to$ 3D & 2 & .28 & .437 & .554 & .099 & .08 & .794 & .374 & .464 & .579 \\
    Text $\to$ 3D, best of 4 & 2 & .06 & .090 & .033 & .107 & .19 & .420 & .403 & .438 & .543 \\
    Text $\to$ 3D, best of 4 & 1 & .04 & .092 & .011 & .126 & .23 & .426 & .395 & .409 & .505 \\
    Ours, best of 4 & 2 & .16 & .327 & .063 & .111 & .23 & .404 & \textbf{.447} & \textbf{.516} & .594 \\
    Ours, best of 4 & 1 & \textbf{.03} & .150 & .025 & .118 & .23 & .413 & .446 & .515 & \textbf{.600} \\
    \midrule
    True labels $\to$ 3D & 2 & .07 & .213 & .223 & .120 & .26 & .412 & .851 & .943 & .876 \\
    Majority class & & & & & & & & .342 & .450 & .598 \\
    \bottomrule
  \end{tabular}}
  \caption{\textbf{Additional geometric metrics and depth sensitivity} (test briefs; $w$: guidance weight; $K$-prior and LLM labels: single samples). Lower is better for Pose FD and Layout W1; contact, feet, and off-screen are best when close to the real shots.}
  \label{tab:s-3d}
\end{table*}

\paragraph{Judging stagings against the story.}
For 300 test briefs (two or more people, one per still, fixed seed) GPT-6.1 Terra receives the brief and an image with the two stagings side by side, ours and Text$\to$3D (best of four, same renderer), with ours on the left in half of the cases, and is asked: \emph{``Choose the staging that better shows the story: who the people are to each other, what is happening between them, and the mood (poses, positions, whom each faces or looks at, distances, framing, light). Ignore rendering quality.''} One answer could not be parsed. Ours is preferred in 58.2\% of the 299 judgments (95\% CI 52.5--63.7, bootstrap over 56 movies); the left image is chosen in 48.8\%.

\paragraph{Expert evaluation.}
Nine expert illustrators evaluated stagings blind. Across the 540 judgments reported in the main paper, our stagings were preferred over Text$\to$3D in 71.1\% of comparisons (95\% CI [58.9, 81.5]) and over retrieval in 72.6\% (95\% CI [59.6, 84.1]); $p\leq 0.002$.

\paragraph{Robustness to brief generation and language-model baselines.}
When Claude Opus 5.5 rewrites the test briefs from the frames alone, without MovieNet captions, our label agreement is .620 [.608, .629], compared with .581 [.568, .592] for Text$\to$3D. When Claude Opus 5.5 instead serves as the language-model label baseline, the resulting staging agreement is .565 [.554, .575].

\paragraph{Images from the staging vs.\ from the brief.}
For 30 held-out briefs (the twelve figure examples and 18 random ones) we compare Seedream images made from the brief alone and from our staging (watercolor). DeepSeek-V4.1-Flash numbers the people from left to right and reports the count, each posture, who looks at and faces whom, and the light, which we compare with the labels of our staging (Table~\ref{tab:s-img}; one brief returned no reading). The quantitative comparison covers these two conditions; the box and skeleton controls are compared qualitatively below. Images made from the staging follow its posture and its light position and color more often; gaze and facing are followed equally often; the key and especially the number of people are followed less often, because the generator brightens dark scenes and drops figures cut by the frame or adds extra ones.

\begin{table}[h]
  \centering
  \small
  \setlength{\tabcolsep}{2.5pt}
  \resizebox{\linewidth}{!}{%
  \begin{tabular}{@{}lcccccccc@{}}
    \toprule
    Image from & count & posture & look & face & light pos. & key & temp. & mean \\
    \midrule
    Brief only & \textbf{.93} & .49 & .60 & .50 & .47 & \textbf{.73} & .57 & .613 \\
    Our staging & .62 & \textbf{.69} & .60 & .51 & \textbf{.69} & .62 & \textbf{.72} & \textbf{.636} \\
    \bottomrule
  \end{tabular}}
  \caption{\textbf{Generated images vs.\ the staging}, read by a vision-language model (29 briefs; 78 postures, 150 ordered pairs).}
  \label{tab:s-img}
\end{table}

\paragraph{Zero-shot on InterHuman.}
We stage the middle frame of each of the 1{,}176 test sequences of InterHuman~\cite{liang2024intergen} (two people, motion capture) from each of its 3{,}528 human-written descriptions, with the trained models unchanged and one sample per description (seed 0, 30 steps, $w=2$). For each ordered pair, the facing angle is the horizontal angle between figure $i$'s body forward direction and the direction to figure $j$, and the head-direction angle the same with the head's forward direction; the error is the absolute difference between generated and captured angles, averaged within each sequence; 95\% CIs bootstrap over sequences (Table~\ref{tab:s-interhuman}). Paired with Text$\to$3D, ours lowers the facing error by $5.8^\circ$ [4.6, 6.9] and the head-direction error by $7.0^\circ$ [6.0, 8.1] ($p<0.001$).

\begin{table}[!ht]
  \centering
  \small
  \setlength{\tabcolsep}{3pt}
  \resizebox{\linewidth}{!}{%
  \begin{tabular}{@{}lcc@{}}
    \toprule
    & Facing error ($^\circ$)$\downarrow$ & Head-direction error ($^\circ$)$\downarrow$ \\
    \midrule
    Random training staging & 48.1 [47.2, 49.2] & 45.7 [44.8, 46.6] \\
    Count only ($K$) & 48.9 [47.8, 49.8] & 46.7 [45.8, 47.6] \\
    Text$\to$3D & 49.0 [47.9, 50.0] & 46.6 [45.6, 47.5] \\
    Ours & \textbf{43.2} [42.1, 44.4] & \textbf{39.6} [38.5, 40.6] \\
    \bottomrule
  \end{tabular}}
  \caption{\textbf{Zero-shot on InterHuman} (1{,}176 test sequences, 3{,}528 descriptions).}
  \label{tab:s-interhuman}
\end{table}

\paragraph{Checking the recovered labels.}
The labels of the data come from single-frame 3D recovery and rules. As an independent check, a vision-language model reads 300 random stills with two or more people (people numbered from left to right) and answers the questions behind each label (Table~\ref{tab:s-frames}). It counts the same number of people in only 54\% of the stills, indicating that its readings are noisy; agreement should therefore be interpreted cautiously. Posture, front/behind, shot size, and key agree clearly above the majority-class rate; light position and temperature slightly; gaze and facing do not, so these two labels should be read as weak supervision.

\begin{table}[h]
  \centering
  \small
  \setlength{\tabcolsep}{2.2pt}
  \resizebox{\linewidth}{!}{%
  \begin{tabular}{@{}lccccccccc@{}}
    \toprule
    & count & posture & look & face & depth & shot & light pos. & key & temp. \\
    \midrule
    Agreement & .54 & .64 & .66 & .54 & .48 & .60 & .44 & .60 & .46 \\
    \quad same count only & -- & .70 & .61 & .50 & .57 & .63 & .44 & .59 & .49 \\
    Majority-class rate & -- & .51 & .78 & .57 & .39 & .39 & .37 & .38 & .36 \\
    \bottomrule
  \end{tabular}}
  \caption{\textbf{Recovered labels vs.\ a vision-language model reading the stills} (300 stills; 756 people, 1{,}270 ordered pairs).}
  \label{tab:s-frames}
\end{table}

\paragraph{Limitations of the measured controls.}
The recovered-label check provides weaker support for gaze and facing than for posture and shot size. Consistent with this, the main paper reports lower editing success for gaze and facing than for shot size and light. Contact is a proximity measure and can also reflect intended physical interaction. The image generator does not always preserve the figure count or the key of the staging. The lighting descriptor captures the appearance of the frame rather than a reconstructed physical light rig. These limitations should be considered when interpreting the qualitative results.

\begin{figure*}[t]
  \centering
  \includegraphics[width=\textwidth]{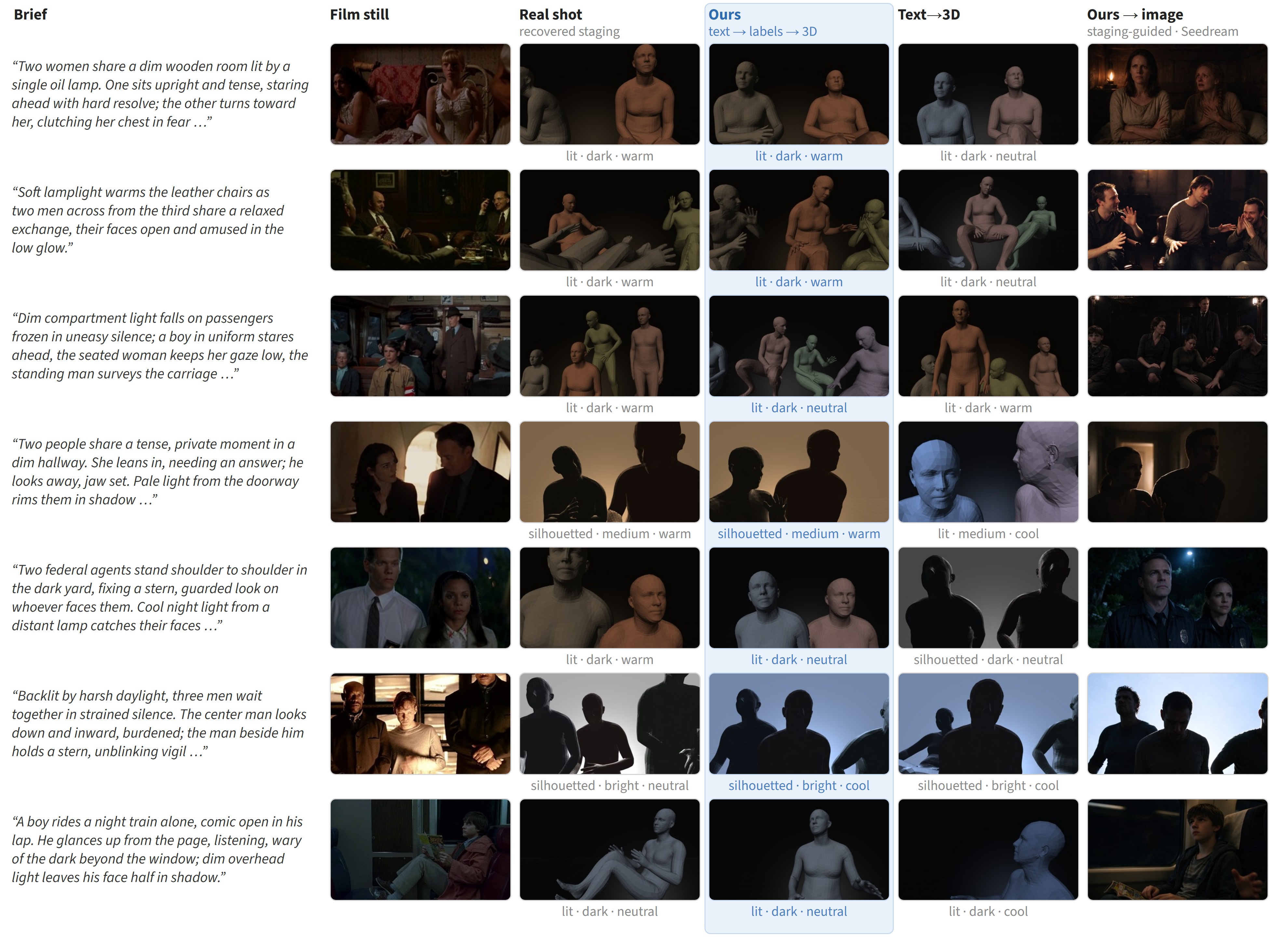}
  \caption{\textbf{All qualitative examples:} film still, real shot, ours, Text$\to$3D (both best of four), and the image generated with our staging (Seedream).}
  \label{fig:s-qual}
\end{figure*}

\begin{figure*}[t]
  \centering
  \includegraphics[width=\textwidth]{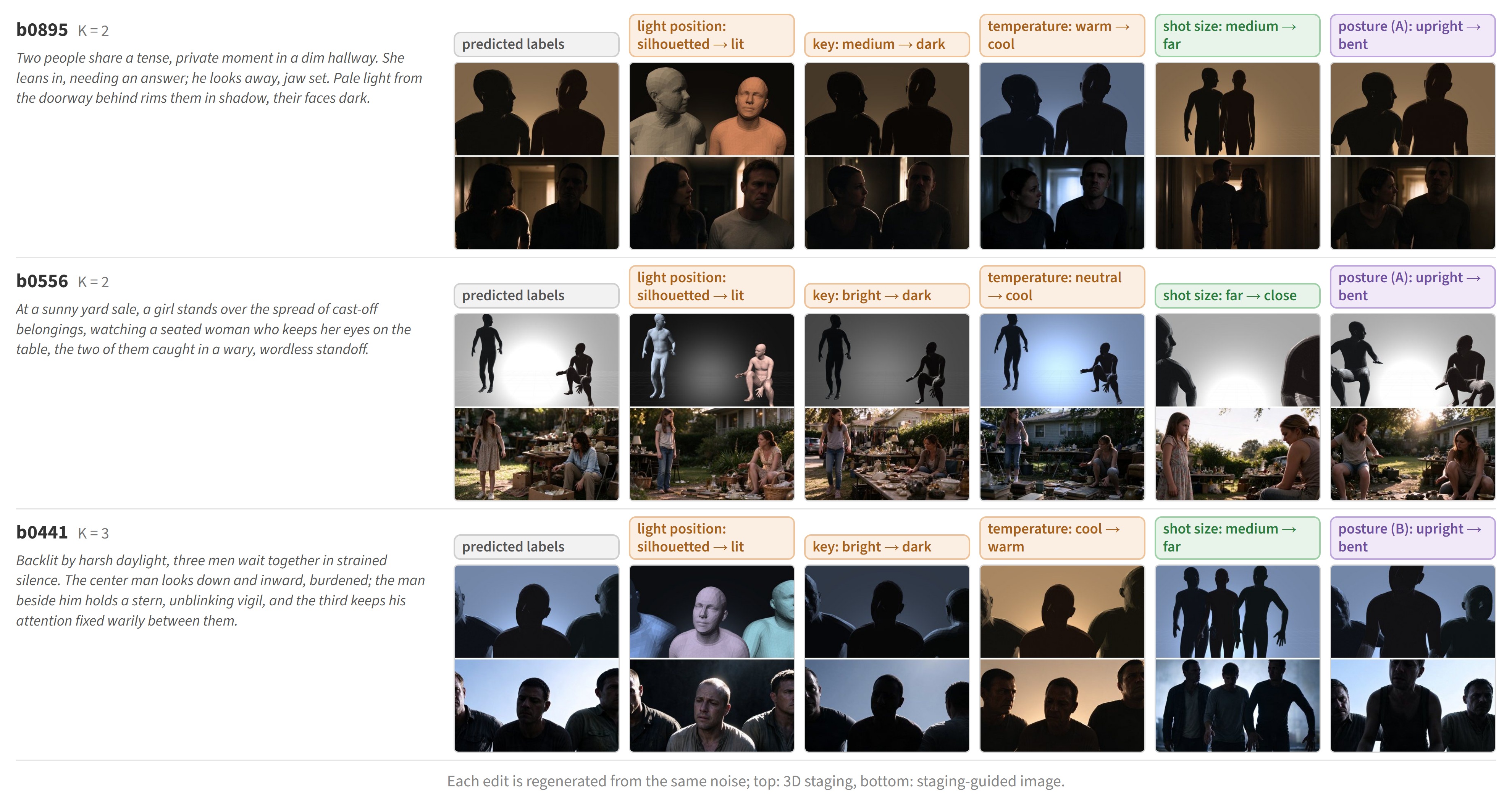}
  \caption{\textbf{Single-label edits} on three briefs. Each staging is regenerated from the same noise with one label changed (top: 3D staging; bottom: the image generated with it).}
  \label{fig:s-edits}
\end{figure*}

\paragraph{More qualitative results.}
Figure~\ref{fig:s-qual} shows seven test briefs, the three of the paper among them, with the image generated from our staging. Figure~\ref{fig:s-edits} shows single-label edits on three briefs; success rates over all test briefs are in Table~\ref{tab:edit} of the main paper.

\begin{figure*}[t]
  \centering
  \includegraphics[width=\textwidth]{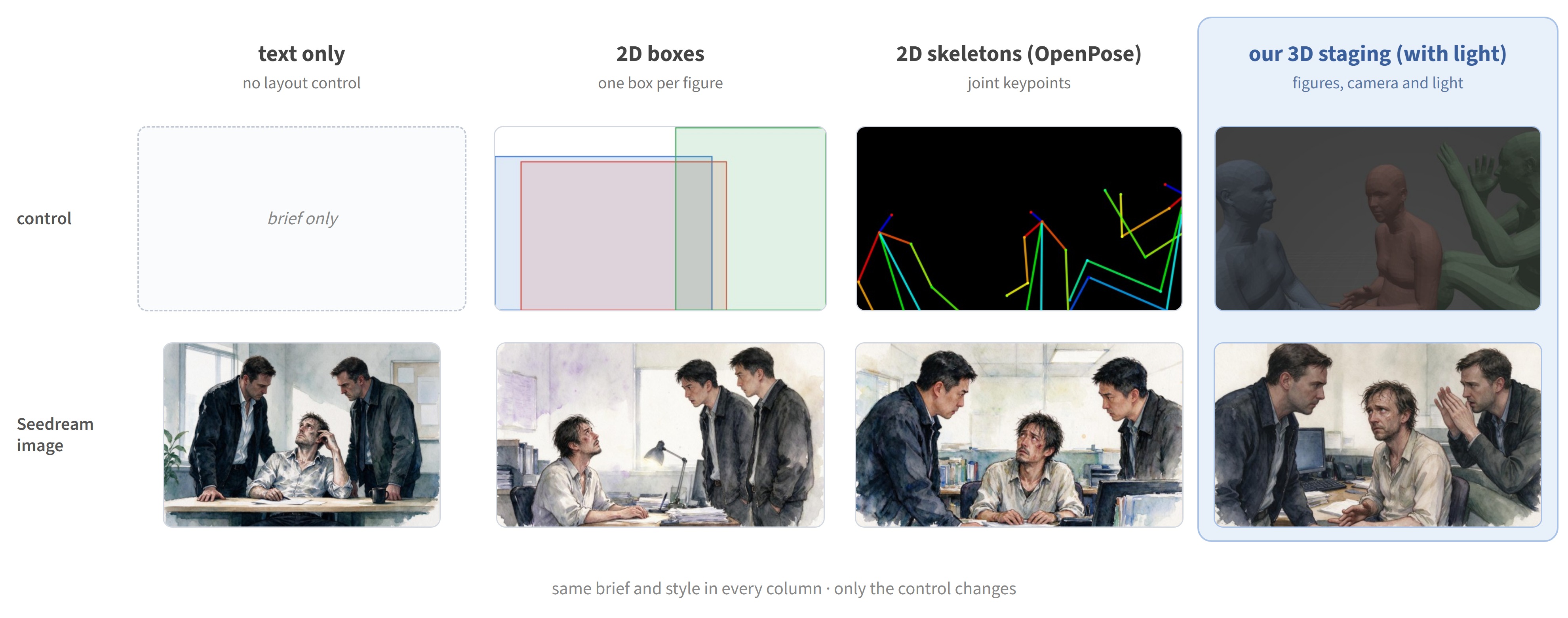}
  \caption{\textbf{The same brief and style with different controls.} Top: controls derived from one generated staging (none, 2D boxes, 2D skeletons, our 3D staging rendered with its light). Bottom: Seedream outputs. With the brief only, two standing men lean in from either side; boxes and skeletons do not fix the arrangement either. The rendered staging transfers its specific choices: the close framing, the seated man's open hand, and the right figure's raised hands. Its light is only partly followed: the flat light is largely retained, but the dark key is not.}
  \label{fig:s-control}
\end{figure*}

\paragraph{Staging as a control for image generation.}
Figure~\ref{fig:s-control} shows the comparison summarized in Sec.~\ref{sec:control-eval} of the main paper. All spatial controls are derived from the same generated staging. Seedream 5.0 Flash receives the same brief and style in all four conditions, including the condition without spatial control.

\begin{figure*}[p]
  \centering
  \includegraphics[width=\textwidth]{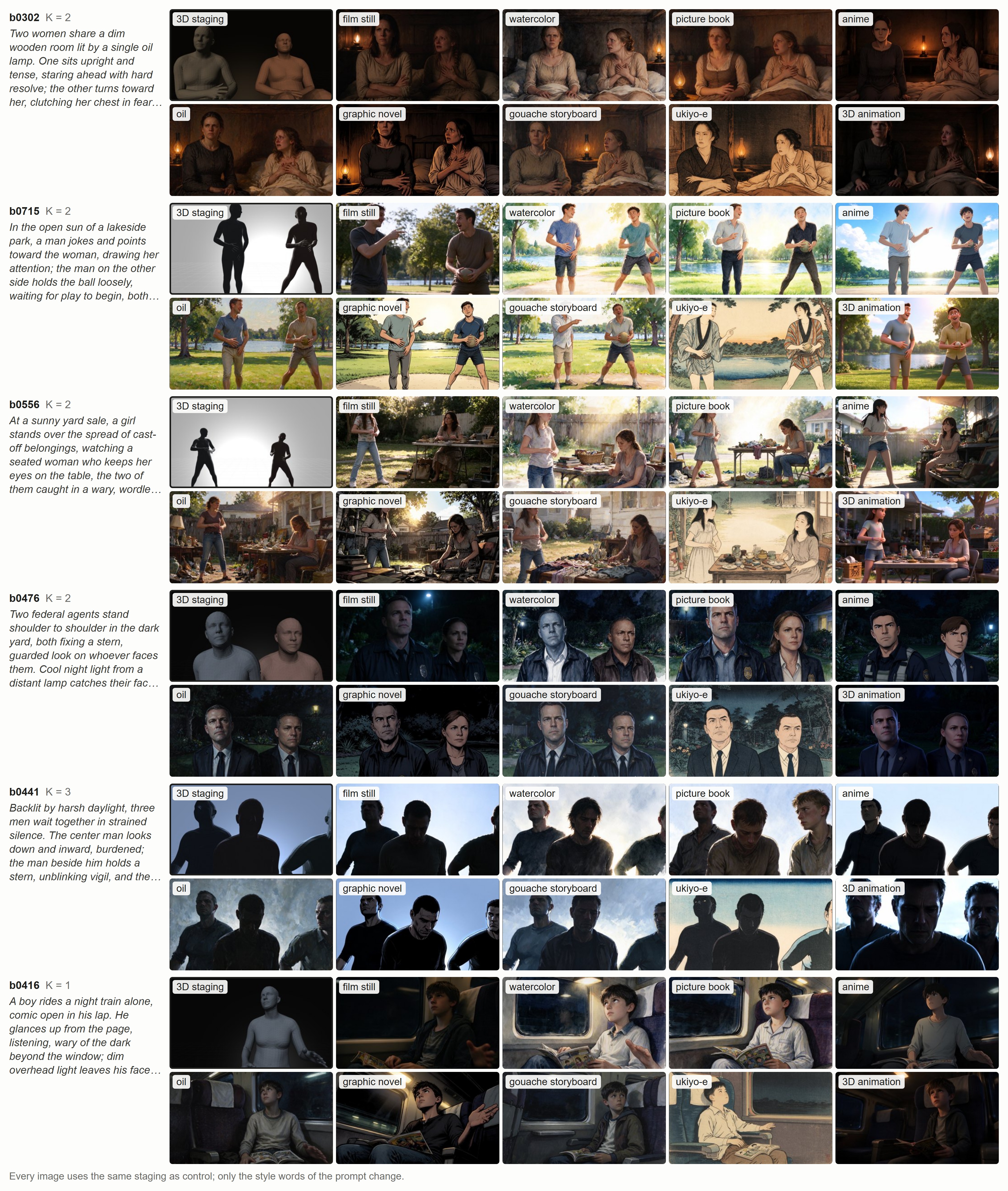}
  \caption{\textbf{One staging, nine styles.} Brief, our staging (text $\to$ labels $\to$ 3D), and the images Seedream generates with it as control.}
  \label{fig:s-gal0}
\end{figure*}

\paragraph{Gallery.}
Figure~\ref{fig:s-gal0} shows six held-out briefs, our staging, and the images Seedream makes from that same staging in nine styles; only the style words of the prompt change. Poses, framing, and light largely carry over across styles, such as the oil lamp, the backlit trio, and the night light. The image prompt asks for natural hairstyles, which avoids the bald heads that the untextured bodies otherwise cause. Briefs for which the image service's content filter refused some styles are left out.

%======================================================================
\section{The ArtStage editor}
\label{sec:editor}

\begin{figure*}[t]
  \centering
  \includegraphics[width=\linewidth]{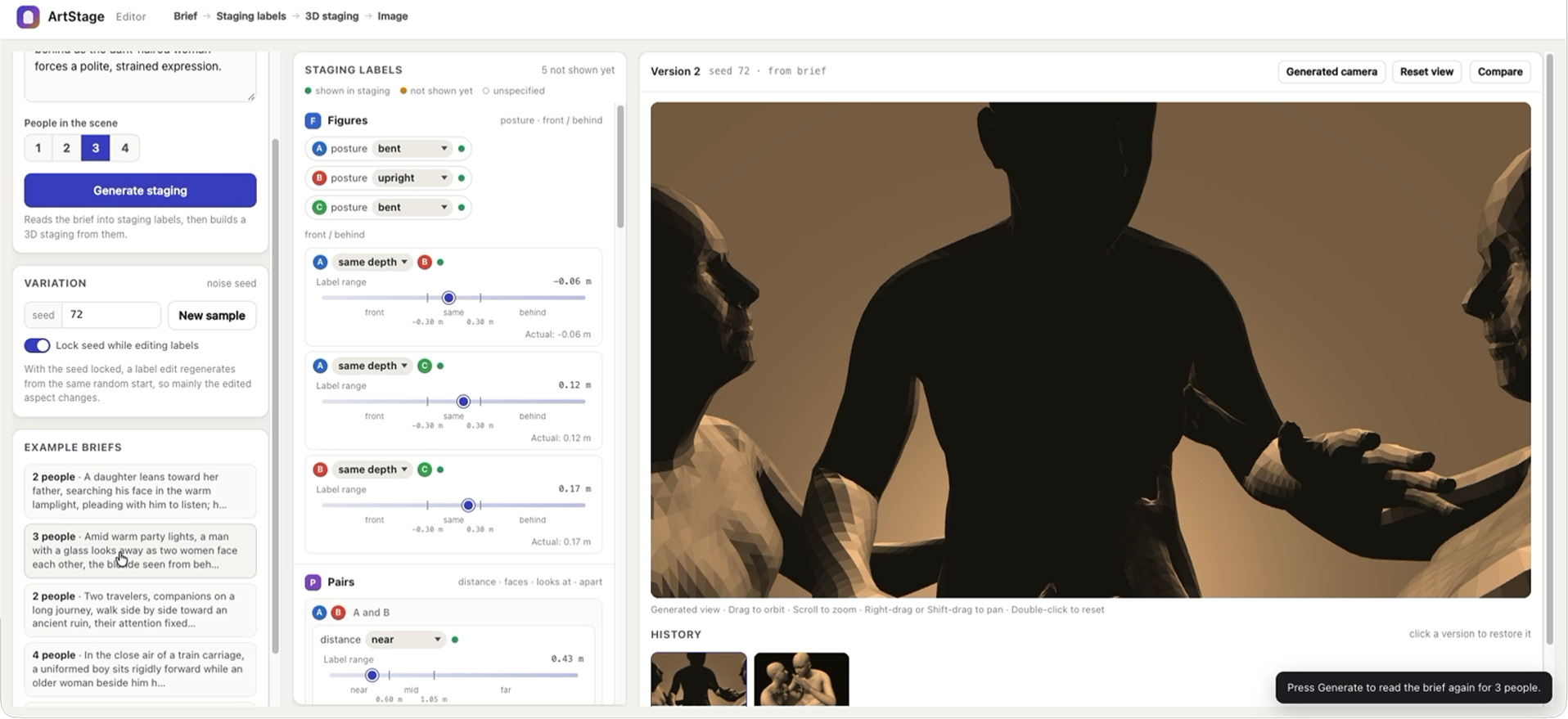}
  \caption{\textbf{The ArtStage editor.} Left: narrative brief, figure count, and sampling seed. Middle: editable staging labels, with sliders marking class boundaries for continuous measurements and status dots indicating label agreement. Right: the interactive 3D viewport with the generated lighting, camera and orbit controls, and version history.}
  \label{fig:cs_interface}
\end{figure*}

Figure~\ref{fig:cs_interface} shows the interactive editor built on the two models. The predicted labels are grouped into Figures, Pairs, Shot, and Light. Each label can be changed through a drop-down menu or left unspecified. With the seed locked, a label edit regenerates the staging from the same noise, so the edited aspect can change while many other labels remain unchanged. The user can also draw a new sample with a different seed and compare or restore earlier versions.

The editor also offers sliders for distance, front/behind, facing, gaze, shot size, figure--ground contrast, key, and color temperature. The sliders show the class boundaries defined in Appendix~\ref{sec:s-data} and are synchronized with the corresponding drop-down menus; shot size is displayed as camera radius in meters rather than its logarithm. Moving a slider across a class boundary selects the new label, and releasing it regenerates the staging if the label changed. The sliders select discrete conditioning labels rather than enforcing exact continuous geometry or lighting values. The measured values of the generated staging are displayed separately. Posture and set apart retain drop-down controls because their definitions involve multiple conditions.

The 3D viewport opens at the generated camera and supports mouse-driven orbit, zoom, and pan, with a reset to the generated viewpoint. These inspection operations do not change the underlying staging or its generated camera. The camera and conventional light direction are shown as guides during inspection. The editor retains version history and exports the staging parameters and the generated-camera rendering; when the optional Seedream integration is configured, that rendering can be used to generate an image in a chosen style.

\end{document}